\documentclass[letterpaper]{article} 
\usepackage{aaai24}  
\usepackage{times}  
\usepackage{helvet}  
\usepackage{courier}  
\usepackage[hyphens]{url}  
\usepackage{graphicx} 
\usepackage{natbib}  
\usepackage{caption} 
\usepackage{algorithm}
\usepackage{booktabs}       
\usepackage{amsfonts}       
\usepackage{nicefrac}       
\usepackage{microtype}      
\usepackage{xcolor}         
\usepackage{graphicx}
\usepackage{amssymb}
\usepackage{appendix}
\usepackage{amsmath}
\usepackage[noend]{algpseudocode}
\usepackage{algorithmicx,algorithm}
\usepackage{appendix}
\usepackage{multirow}
\usepackage{mathrsfs}
\usepackage{amsthm}
\usepackage{float}
\usepackage{dsfont}
\usepackage[misc]{ifsym}
\usepackage{diagbox}

\DeclareMathOperator*{\argmin}{arg\,min}
\DeclareMathOperator*{\argmax}{arg\,max}

\usepackage{newfloat}
\usepackage{listings}
\DeclareCaptionStyle{ruled}{labelfont=normalfont,labelsep=colon,strut=off} 
\floatstyle{ruled}
\newfloat{listing}{tb}{lst}{}
\floatname{listing}{Listing}
\title{Compositional Text-to-Image Synthesis with Attention Map Control of Diffusion Models}
\author{
    Ruichen Wang\textsuperscript{\rm 1},
    Zekang Chen\textsuperscript{\rm 2}\thanks{The author did his work during internship at OPPO Research Institute. \# denotes corresponding authors.}, 
    Chen Chen\textsuperscript{\rm 1 \#}, 
    Jian Ma\textsuperscript{\rm 1},
    Haonan Lu\textsuperscript{\rm 1 \#}, 
    Xiaodong Lin\textsuperscript{\rm 3}
}
\affiliations{
    \textsuperscript{\rm 1}OPPO Research Institute, 
    \textsuperscript{\rm 2}South China University of Technology,
    \textsuperscript{\rm 3}Rutgers University \\

    wangruichen@oppo.com,chenzekang2018@163.com,\{chenchen4,majian2,luhaonan\}@oppo.com, lin@business.rutgers.edu\\
}

\usepackage{bibentry}

\begin{document}

\maketitle

\begin{abstract}
Recent text-to-image (T2I) diffusion models show outstanding performance in generating high-quality images conditioned on textual prompts. However, they fail to semantically align the generated images with the prompts due to their limited compositional capabilities, leading to attribute leakage, entity leakage, and missing entities. In this paper, we propose a novel attention mask control strategy based on predicted object boxes to address these issues. In particular, we first train a BoxNet to predict a box for each entity that possesses the attribute specified in the prompt. Then, depending on the predicted boxes, a unique mask control is applied to the cross- and self-attention maps. Our approach produces a more semantically accurate synthesis by constraining the attention regions of each token in the prompt to the image. In addition, the proposed method is straightforward and effective and can be readily integrated into existing cross-attention-based T2I generators. We compare our approach to competing methods and demonstrate that it can faithfully convey the semantics of the original text to the generated content and achieve high availability as a ready-to-use plugin. Please refer to  https://github.com/OPPO-Mente-Lab/attention-mask-control.
\end{abstract}

\begin{figure}[htbp]
	\centering
	\includegraphics[width=.46\textwidth]{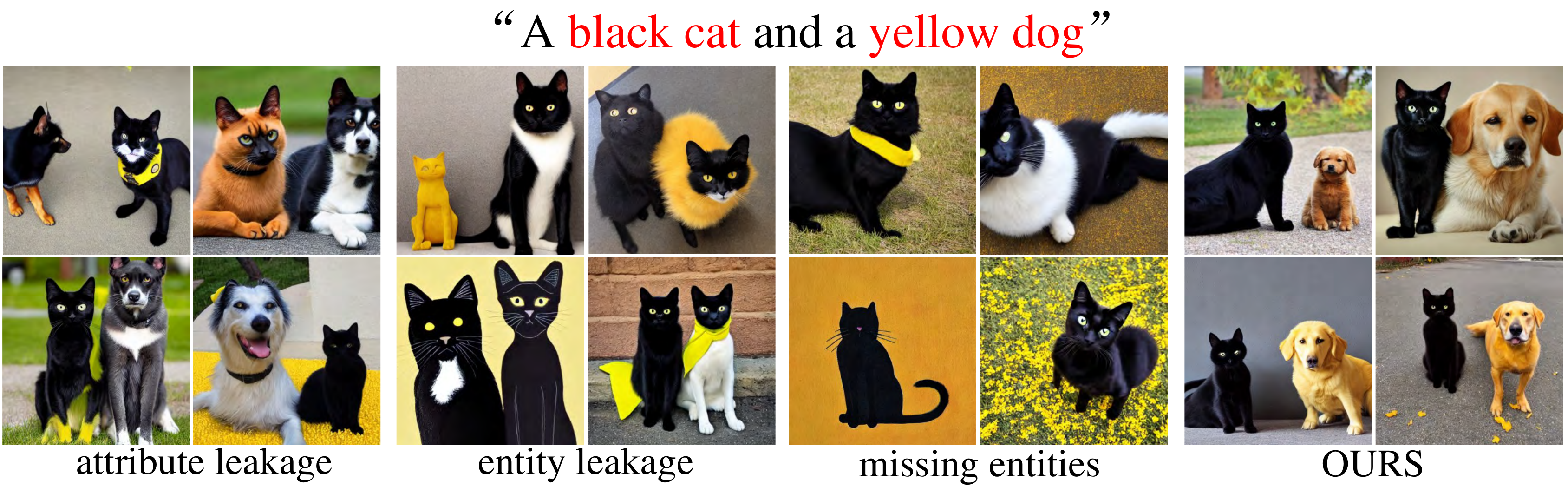}
	\caption{Example results from Stable Diffusion (first three sets of images) and Our method (last set). Our method aims to address three typical generation defects (\textit{attribute leakage}, \textit{entity leakage}, and \textit{missing entities}) and generate images that are more semantically faithful to the image captions.}
	\label{examples}
\end{figure}

\section{Introduction}
Text-to-image (T2I) synthesis aims to generate realistic and diverse images conditioned on text prompts. Recently, diffusion models have achieved state-of-the-art results in this area~\cite{rombach2022high,croitoru2023diffusion,yang2022diffusion}. Compared to previous generative models, such as generative adversarial networks (GANs)~\cite{goodfellow2020generative} and variational autoencoder (VAE)~\cite{doersch2016tutorial}, diffusion models exhibit superior performance with respect to image generation quality and diversity. They also enable better content control based on the input conditions such as grounding boxes, edge maps, or reference images, while avoiding the problems of training instability and mode collapse~\cite{zhang2023adding,li2023gligen}.

Despite their success, diffusion-model-based synthesis methods struggle to accurately interpret compositional text descriptions, especially those containing multiple objects or attributes \cite{feng2022training,han2023svdiff,liu2023cones,chefer2023attend,jimenez2023mixture}. The generation defects of diffusion models such as Stable Diffusion (SD)~\cite{rombach2022high}fall into three categories: attribute leakage, entity leakage, and missing entities, as shown in Fig.\ref{examples}. Considering the prompt ``a black cat and a yellow dog'', attribute leakage refers to the phenomenon where the attribute of one entity is observed in another (\textit{e.g.,} a black dog). Entity leakage occurs when one entity overlays another (\textit{e.g.,} two cats, one black and one yellow). Missing entities indicate that the model fails to generate one or more of the subjects mentioned in the input prompt (\textit{e.g.,} only one black cat).

We attribute the infidelity issues in T2I synthesis to inaccurate attention regions, \textit{i.e.,} the cross-attention regions between text tokens and image patches, as well as the self-attention regions within image patches themselves. Each entity and its attribute should, ideally, correspond to a coherent image region in order to generate multiple entities in a single image correctly. Existing T2I diffusion models, such as SD, lack explicit constraints on the attention regions and boundaries, which may lead to overlapping attention activations. To address these issues, we attempt to use parsed entities with attributes and their predicted object boxes to provide explicit attention boundary constraints for compositional generations. Specifically, predicted object boxes define the interest areas on images, while entities with attributes depict the interest text spans where each text token shares a common cross-attention region. By incorporating these boundary constraints, we achieve high-fidelity T2I synthesis while addressing the aforementioned problems.

In this paper, we propose a novel compositional T2I approach based on SD~\cite{rombach2022high} with explicit control of cross- and self-attention maps to ensure that the attention interest areas are located within the predicted object boxes, as shown in Fig.\ref{model_overview}. Specifically, we first train a BoxNet applied to the forward process of SD on the COCO dataset~\cite{lin2014microsoft} to predict object boxes for entities with attributes parsed by a constituency parser~\cite{honnibal2020spacy}. 
We then enforce unique attention mask control over the cross- and self-attention maps based on the predicted boxes (image regions) and entities with attributes (text spans).
The objective of BoxNet is to provide entity-bounding boxes for subsequent attention mask control. During the diffusion model's inference process, random sampling is performed at each step, so the box positions of each entity are constantly changing (as shown in Fig.\ref{box}). Consequently, the BoxNet predicts the entity box positions at each step based on the diffusion model's intermediate features to avoid excessive conflict between the current sampling result and the mask control target.
Our approach produces a more semantically accurate synthesis by constraining the attention region of each text token in the image. Furthermore, using the trained BoxNet, our method can guide the diffusion inference process on the fly without fine-tuning SD. We conduct comprehensive experiments on the publicly available COCO and open-domain datasets, and the results show that our method generates images that are more closely aligned with the given descriptions, thereby improving fidelity and faithfulness.
The main contributions of our work can be concluded as follows:
\begin{itemize}
  \item We propose BoxNet, an object box prediction module capable of estimating object locations at any timestep during the forward diffusion process. The predicted object boxes closely match the locations of the entities generated by the original SD.
  \item We develop an effective attention mask control strategy based on the proposed BoxNet, which constrains the attention areas to lie within the predicted boxes.
  \item The trained BoxNet and attention mask control of our method can be easily incorporated into existing diffusion-based generators as a ready-to-use plugin. We demonstrate our model's capability by integrating it into the original SD and two variants: Attend-and-Excite~\cite{chefer2023attend} and GLIGEN~\cite{li2023gligen}.
\end{itemize}

\section{Related Work}
\textbf{Text-to-Image Diffusion Models.} Diffusion models are becoming increasingly popular in the T2I synthesis area due to their exceptional performance in generating high-quality images~\cite{ramesh2021zero,esser2021taming,ramesh2022hierarchical,balaji2022ediffi,saharia2022photorealistic}. Generally, these models take a noisy image as input and iteratively denoise it back to a clean one while semantically aligning the generated content with a text prompt. SD~\cite{rombach2022high} uses an autoencoder to create a lower-dimensional space and trains a U-Net model~\cite{ronneberger2015u} based on large-scale image-text datasets in this latent space, balancing algorithm efficiency and image quality. However, diffusion models have limited expressiveness, resulting in generated content that cannot fully convey the semantics of the original text. This issue is exacerbated when dealing with complex scene descriptions or multi-object generation~\cite{chefer2023attend,feng2022training,ma2023directed}.

\textbf{Compositional Generation.} Recent studies have explored various approaches to enhance the compositional generation capacity of T2I diffusion models without relying on additional bounding box input. StructureDiffusion~\cite{feng2022training} uses linguistic structures to help guide image-text cross-attention. However, the results it produces frequently fall short of addressing semantic issues at the sample level. Composable Diffusion~\cite{liu2022compositional} breaks down complex text descriptions into multiple easily-generated snippets. A unified image is generated by composing the output of these snippets. Yet, this approach is limited to conjunction and negation operators. AAE~\cite{chefer2023attend} guides a pre-trained diffusion model to generate all subjects mentioned in the text prompt by strengthening their activations on the fly. Although AAE can address the issue of missing entities, it still struggles with attribute leakage and may produce less realistic images when presented with an atypical scene description.
Unlike previous methods, our work proposes a novel two-phase method of BoxNet and Attention Mask Control, gradually controlling the generation of multiple entities during the diffusion model sampling process.

\textbf{Layout to Image Generation.} Through the use of artificial input conditions such as bounding boxes, shape maps, or spatial layouts, some existing methods can generate controllable images.
For instance, GLIGEN~\cite{li2023gligen} adds trainable gated self-attention layers to integrate additional inputs, such as bounding boxes, while freezing the original model weights. Chen \textit{et al.}~\cite{chen2023training} propose a training-free layout guidance technique for guiding the spatial layout of generated images based on bounding boxes. Shape-Guided Diffusion~\cite{huk2022shape} leverages an inside-outside attention mechanism during the generation process to apply the shape constraint to the attention maps based on a shape map.
However, these works require prior layout information to be provided as input, which is fixed during the generation process. In order to directly control the generation results of diffusion models, our work aims to provide a pure text-to-image generation method that does not require users to specify bounding boxes or layouts. Instead, BoxNet estimates such information at each sampling step.

\textbf{Layout-based Generation.} Some work can directly generate the layout based on user input text information and further generate images based on layouts.
Wu \textit{et al.}~\cite{wu2023harnessing} address the infidelity issues by imposing spatial-temporal attention control based on the pixel regions of each object predicted by a LayoutTransformer~\cite{yang2021layouttransformer}. However, their algorithm is time-consuming, with each generation taking around 10 minutes. Also, Lian \textit{et al.}~\cite{lian2023llm} propose to equip diffusion models with off-the-shelf pretrained large language models (LLMs) to enhance their prompt reasoning capabilities. But this approach is highly dependent on LLMs, which are hard to control and prohibitively expensive to deploy.
Current layout-based approaches typically split image generation into two completely disrelated stages: prompt-to-layout and layout-to-image, while our method optimizes the diffusion model itself by performing step-wise box prediction and generation control at each sampling step to maintain the original capability of the model while improving the entity properties.

\section{Method}
\begin{figure}[tbh]
	\centering
	\includegraphics[scale=0.21]{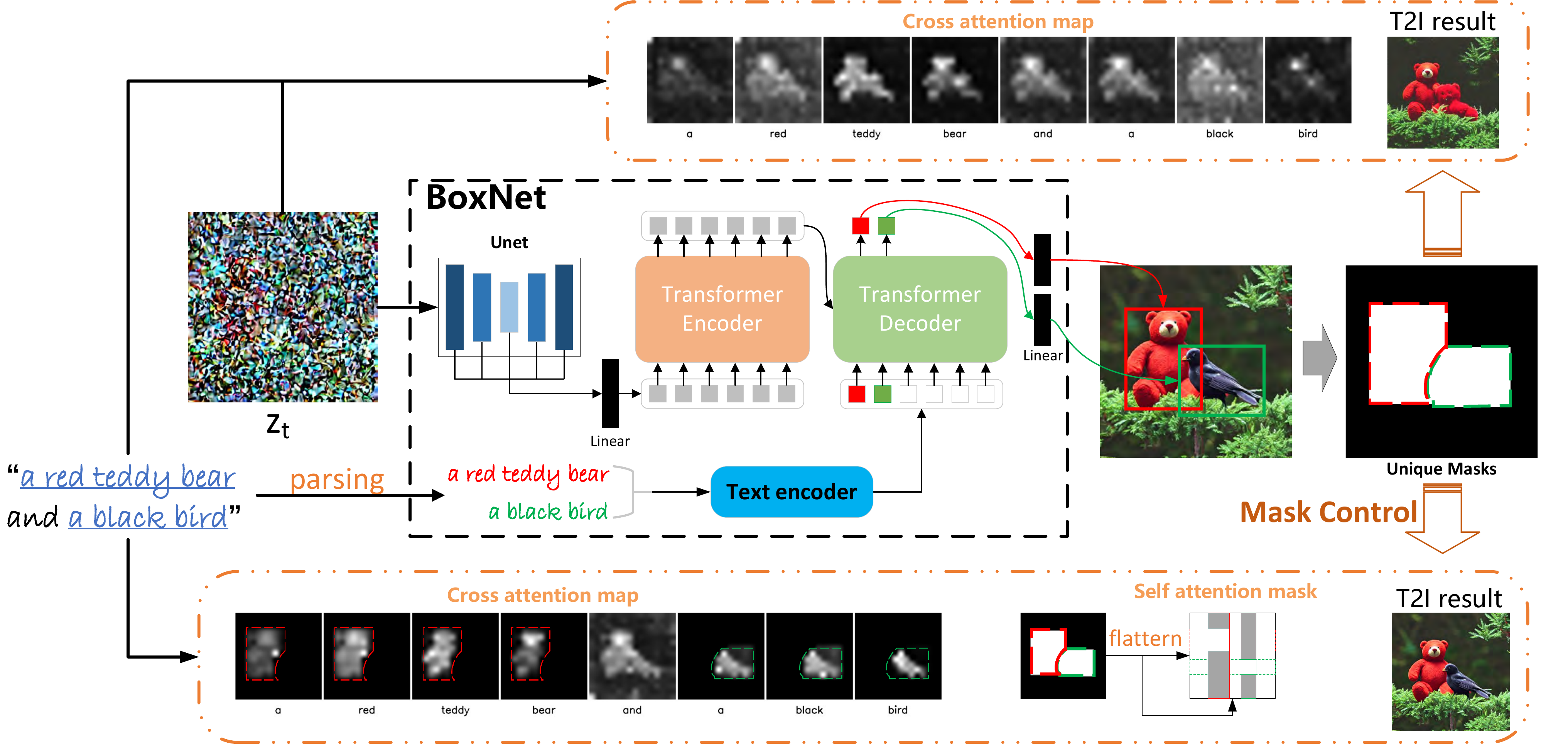}
	\caption{Overview of our BoxNet-based T2I generation pipeline. BoxNet consists of a text encoder and a U-Net followed by an encoder-decoder transformer, as shown in black dashed box. BoxNet takes as input a text prompt, a noisy image, and a timestep and outputs boxes that specify objects' locations. The orange dashed box shows the attention mask control strategy enforced over the cross-attention maps conditioned on the boxes (image regions) and phrases (text spans) as well as the self-attention maps.}
	\label{model_overview}
\end{figure}
Algorithm~\ref{pseudo_code} shows the overall pipeline of our method, which contains two main parts: BoxNet that predicts a box for each entity with attributes, and attention mask control that ensures the generation of accurate entities and attributes. A single denoising step of our model is illustrated in Fig.\ref{model_overview}, in which we use BoxNet to predict the bounding box for each entity parsed from the input text and obtain unique masks via the method in Sec.~\ref{boxnet}. We then perform explicit unique mask control over cross- and self-attention maps on each attention layer of the SD~\cite{rombach2022high}, as explained in Sec.~\ref{attention_control}, which enables to generate entities with their attributes inside the unique mask areas.

The U-Net~\cite{ronneberger2015u} denoiser contains both cross- and self-attention layers. Each cross-attention layer generates a spatial attention map that indicates the image region to which each textual token is paying attention. Similarly, each self-attention layer produces a spatial attention map that represents the interdependence of each patch and all patches. We assume the aforementioned infidelity problems are caused by the inaccurate cross- and self-attention regions in the U-Net. To alleviate the issues, we enforce an attention mask control strategy over attention maps based on the BoxNet during the diffusion backward process, as shown in Fig.\ref{model_overview}. In the original SD, attention regions for the entities ``bear'' and ``bird'' overlap, with the attention of ``bird'' being significantly weaker than that of ``bear'', leading to entity leakage (\textit{i.e.,} generation of two bears). However, after using our method, the prompt ``a red teddy bear is sitting next to a black bird'' is generated correctly.
\begin{algorithm}[tbh]
    \small
	\caption{Denoising Process of Our Method}
	\label{pseudo_code}
    {\bf Input:} A text prompt $p$, a trained BoxNet $B$, sets of each parsed entity's token indices $\{s_1, s_2, ..., s_N\}$, a trained
diffusion model $SD$\\
	{\bf Output:} Denoised latent $z_0$.
	\begin{algorithmic}[1]
        \For{$t \leftarrow T, T-1, ... , 1$}
        \State $boxes \leftarrow B(SD, z_{t}, p, t)$
        \For{($cx,cy,h,w$) in boxes}
        \State Convert box to zero-one masks $m_n$
        \State $G_n \leftarrow Gaussian\_distribution\_2D((cx,cy),h,w)$
        \EndFor
        \State $M \leftarrow argmax(G_n)$
        \State $m_n' \leftarrow (M=n) \odot m_n, \ \  n=1,2...,N$ \Comment{unique masks}
        \State $SD' \leftarrow SD$
        \For{each cross attention layer in $SD'$}  \Comment{cross attention mask control}
        \State Obtain Cross Attention Map $C$
        \State $C_i \leftarrow C_i \odot m_n' \ \ \ \forall \ i \in s_n, n=1,2...,N$ 
        \EndFor
        \For{each self attention layer in $SD'$}  \Comment{self attention mask control}
        \State Obtain Self Attention Map $S$
        \State $S_i \leftarrow S_i \odot flatten(m_n') \ \ \ \forall \ i \in \{ i|flatten(m_n')_i=1 \}, n=1,2...,N$ 
        \EndFor
        \State $z_{t-1} \leftarrow SD'(z_{t}, p, t)$
        \EndFor
	\end{algorithmic}
\end{algorithm}
\subsection{BoxNet Architecture}
\label{boxnet}
Our BoxNet consists of a U-Net feature extractor, a text encoder, and an encoder-decoder transformer~\cite{carion2020end} as shown in Fig.\ref{model_overview}. When training the BoxNet, the U-Net and the text encoder are initialized and frozen from a pretrained SD checkpoint. At each timestep $t$ of the SD denoising process, the U-Net takes as input a noisy image $z_t$, a text prompt $p$, and a timestep $t$, and then we extract the output feature maps from each down- and up-sampling layer of the U-Net. All the extracted feature maps are interpolated to the same size and concatenated together. A linear transformation is then applied to acquire a feature tensor $f$ that represents the current denoised latent $z_t$.

After that, we use a standard encoder-decoder transformer to generate entity boxes. Note that the encoder expects a sequence as input; hence, we reshape the spatial dimensions of $f$ into one dimension, refer to~\cite{carion2020end}.
The decoder decodes boxes with input entity queries. To acquire entity queries, the text prompt input by a user is first parsed into $N$ entities with attributes manually or by an existing text parser~\cite{honnibal2020spacy}, as shown in Fig.\ref{model_overview}. Then, the entity phrases are encoded into embeddings by the text encoder. Entity embeddings are pad with a trainable placeholder tensor into a max length of $M$, and only the first $N$ of the output sequences are used to calculate entity boxes by a weighted shared linear projection layer. 
 
As to the training phase, we train the BoxNet in the forward process of SD on the COCO dataset. It's worth noting that the primary goal of our BoxNet is to assign each entity a reasonable bounding box during each step of generation, which can improve the attention map control to modify entity generation throughout the whole process. We don't concern much about achieving high object recognition accuracy, which differs from DETR. Since one input image may have multiple instance-level ground-truth boxes in the same category, it is necessary to define a proper loss function to constrain our predicted boxes with ground-truth. Inspired by ~\cite{carion2020end}, we first produce an optimal bipartite matching between predicted and ground-truth boxes, and then we optimize entity box losses.
Let us denote by $b$ the ground-truth set of $N$ objects, and $b'$ the set of top $N$ predictions.
To find a bipartite matching between these two sets, we search for a permutation of $N$ elements $\sigma \in P_N$ with the lowest cost:
\begin{equation}
    \hat{\sigma} = \argmin_{\sigma \in P_N} \sum_i^N \mathcal{L}_\text{match}(b_i,b'_{\sigma(i)}),
\end{equation}
where $\mathcal{L}_\text{match}(b_i,b'_{\sigma(i)}$) is a pair-wise matching cost. This optimal assignment is computed efficiently with the Hungarian algorithm, following prior works~\cite{carion2020end,stewart2016end}. Different from~\cite{carion2020end}, since our BoxNet aims to assign a reasonable bounding box to each object, a precise bounding box with a mismatched category is meaningless. Therefore, we prioritize classification accuracy over location accuracy by modifying the matching cost to include an extremely high penalty for bounding boxes with class mismatches: 
\begin{equation}
    \mathcal{L}_\text{match}(b_i,b'_{\sigma(i)}) =  \lambda \cdot \mathds{1}{\{c_i \neq c_{\sigma(i)}\}} + \mathcal{L}_\text{box}(b_i,b'_{\sigma(i)})
\end{equation}
where $c_i$ is the target class label, $c_{\sigma(i)}$ the predicted class label, and $\mathcal{L}_\text{box}(\cdot,\cdot)$ the entity box loss described below. We assign $\lambda$ an extremely high value to avoid class mismatches.
The next step is to compute the loss function of BoxNet. We use a linear combination of the $L1$ loss and the generalized IoU loss $\mathcal{L}_\text{box}(\cdot,\cdot)$ from \cite{rezatofighi2019generalized}.
\begin{equation}
    \mathcal{L}_\text{box}(b_i,b'_{\hat{\sigma}(i)}) = \lambda_{iou}\mathcal{L}_\text{iou}(b_i,b'_{\hat{\sigma}(i)}) + \lambda_{L1}\left| b_i-b'_{\hat{\sigma}(i)} \right|
\end{equation}
where $\lambda_{iou}$, $\lambda_{L1}$ are hyperparameters.

\begin{figure}[tb]
	\centering
	\includegraphics[scale=0.35]{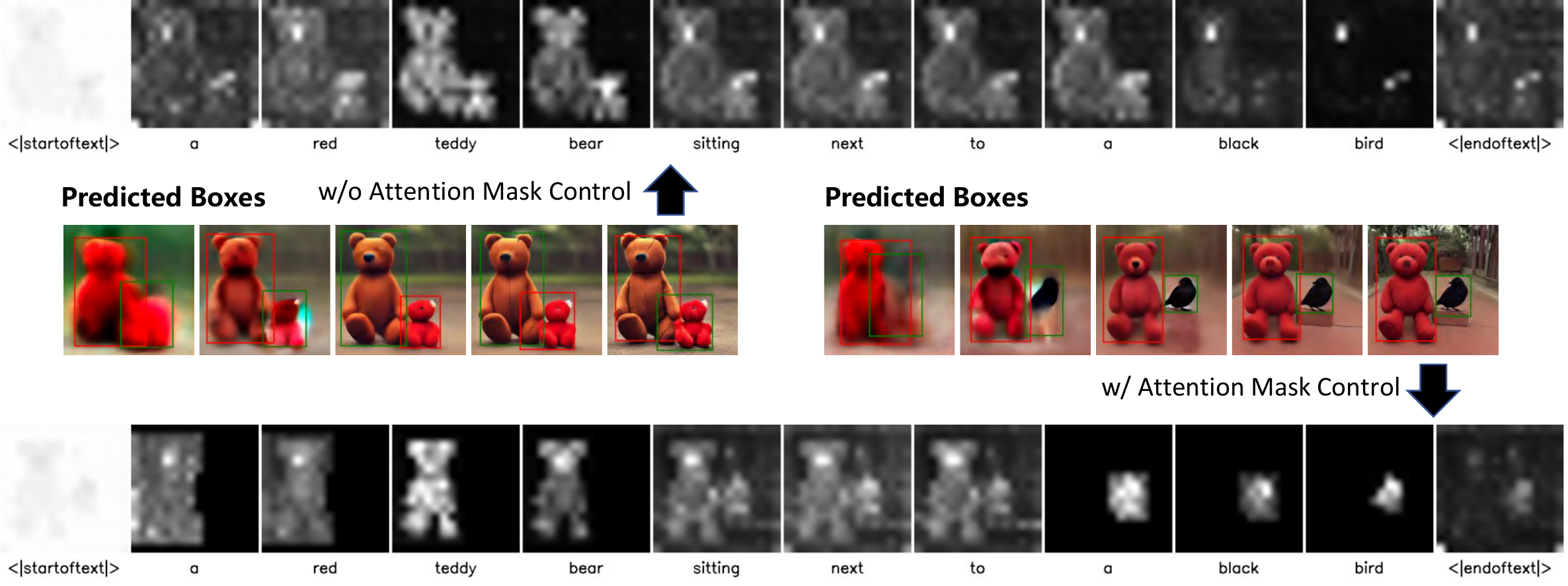}
	\caption{BoxNet predicted box results and corresponding cross-attention maps are presented. We can generate better multi-entity images by controlling the attention map.}
	\label{box}
\end{figure}
Though the BoxNet is trained on the COCO dataset with finite entity classification, we observe that it can also generalize well to unseen entities beyond the COCO dataset (NON-COCO dataset), which implies that the BoxNet, trained on the COCO dataset, establishes a mapping relationship between entity name embeddings and the SD model's intermediate generation results.
In addition, as shown in Fig.\ref{box}, the prediction results of the BoxNet match the location of entities with attributes generated by the original diffusion model even when infidelity problems occur. This provides us with the possibility to control the interest area of each entity on attention maps through predicted boxes.

\subsection{Attention Mask Control}
\label{attention_control}
Before performing attention mask control, the predicted boxes need to be converted into zero-one masks. However, for those entity boxes with severe overlap, it is hard to limit each entity to its own area of interest, which may degrade the multi-entity controllability. So we introduce a unique mask algorithm that generates unique zero-one masks for attention map control. This ensures that each entity has its own area of interest and does not interfere with each other. Since self-attention maps heavily influence how pixels are grouped to form coherent entities, we also apply a similar manipulation to them based on the masks to further ensure that the desired entities and attributes are generated.

\textbf{Unique Mask Algorithm.} Assume we have predicted entity boxes, and they are converted to zero-one masks $m_n$, $n={1,2,...,N}$. For each entity box $(c_x, c_y, w, h)$, we employ an independent 2-dimensional Gaussian distribution probability function $G_n$ with two variances $\nu_1=w/2$ and $\nu_2=h/2$, where $c_x, c_y$ means the center coordinate of the box and $w, h$ means the width and height of the box.
\begin{equation}
    G_n(x, y) = \frac{1}{\sqrt{2\pi\nu_1 \nu_2 }}\exp{\left[-\frac{1}{2} \left( \frac{\left(x-c_x \right)^2}{\nu_1} + \frac{\left(y-c_y \right)^2}{\nu_2} \right)\right]}
\end{equation}
$x =1,2,...,W; y=1,2,...,H$ where $W, H$ represent the spatial width and height of attention maps.
Then we can get the max index map $M$ by
\begin{equation}
    M(x, y) = \argmax_{i = 1,2,...,N} \left(G_i(x, y) \right)
\end{equation}
The unique attention masks can be further computed with:
\begin{equation}
\label{eq5}
m_n'(x,y) = \mathds{1}(M(x,y)=n) \odot m_n(x,y), \ \  n=1,2...,N
\end{equation}
Assume we have unique attention masks $m_n'$ with shape $(H, W)$ from Eq.~\ref{eq5}, where $n=1,2...,N$ indicates the unique mask of the $n$-th entity.

\section{Experiments}
\label{experiments}
\begin{figure*}[htbp]
	\centering
	\includegraphics[scale=0.28]{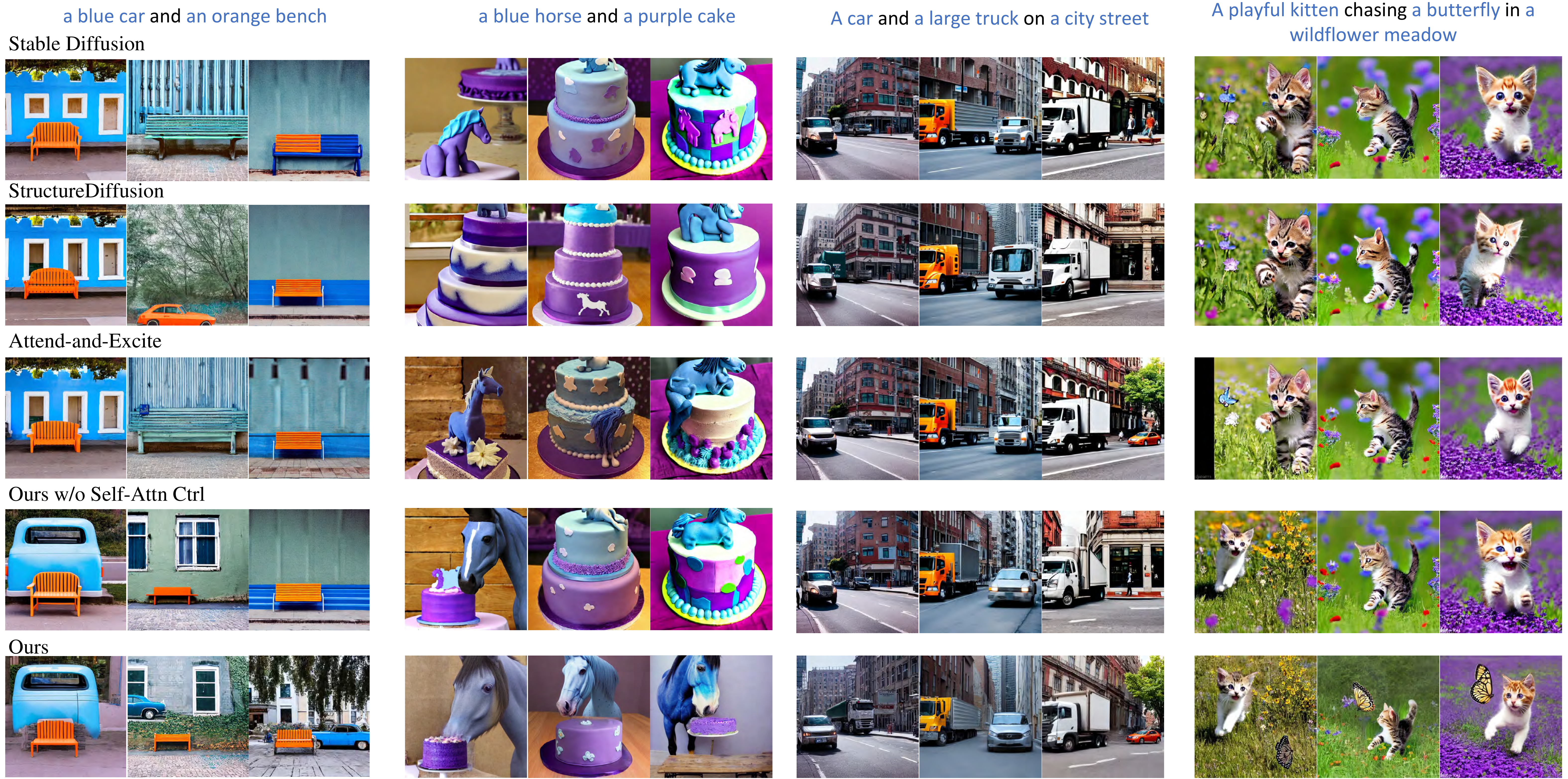}
	\caption{Qualitative comparison of self-built prompts in fixed format (first three columns) and complex prompts in COCO-style (last two columns) with more than two entities and complex attributes. We display four images generated by each of the five competing methods for each prompt, with fixed random seeds used across all approaches. The entities with attributes are highlighted in blue.}
	\label{array_image_qualitative}
\end{figure*}

\textbf{Cross attention mask control.} 
For cross attention, we get the attention map $C$ by:
\begin{equation}
\label{eq7}
    C = softmax\left( \frac{QK^T}{\sqrt{d}} \right)
\end{equation}
In cross-attention, $Q$ comes from the flattened feature map of the model (the Unet of Stable Diffusion); assume the shape is $(L,C)$, where $L=H*W$. While $K$ has the shape $(K,C)$, which represents the embedding of the input prompt $p$, $K$ here is the token number. $\frac{1}{\sqrt{d}}$ is the scaling factor. The self-attention map $C$ has shape $(L,K)$.

For each $n$-th entity, its token indices in the tokenization of $p$ is $s_n$. We can aplly cross attention map control on the cross attention map $C$ by:
\begin{equation}
    C[:,i] = C[:,i] \odot flatten(m_n') \ \ \ \forall \ i \in s_n
\end{equation}

\textbf{Self attention mask control.} 
As the Self Attention Mask Control method. First of all, we get the self attention map $S$ as the same equation in eq.~\ref{eq7}. Differently, in self-attention, both $Q$ and $K$ are from the flattened feature map with shape $(L,C)$. And the attention map $S$ has shape $(L*L)$.
For each $n$-th entity, similar to cross-attention control, we need to choose indices of attention map to be masked. Since $K$ represents the feature map itself, we use the unique mask $m_n'$ to seek indices to be masked instead of $s_n$ in cross-attention mask control. As Fig.\ref{self_attn_ctrl} in Appendix~\ref{appendix:self-attention_mask_control} shows, the self-attention map can be controlled by:
\begin{equation}
    S[:,i] = S[:,i] \odot flatten(m_n') \ \ \ \forall \ i \in \{ i|flatten(m_n')_i=1 \}
\end{equation}

\subsection{Plugin Method}
Once the BoxNet is trained, our method can act as a plugin to guide the inference process of diffusion-based models on the fly, improving the quality of multi-entity generation with attributes. Our BoxNet can provide input conditions for some layout-based generation models, reducing user input and optimizing the efficiency of large-scale data generation. Furthermore, the attention mask control based on predicted boxes can also be directly applied to other T2I generators to address the three infidelity issues. We introduce two plugin solutions using existing models as examples and compare their results with and without our method. For more details, refer to Table~\ref{line_plot_cropped}.

\textbf{AAE}~\cite{chefer2023attend} guides the latent at each denoising timestep and encourages the model to attend to all subject tokens and strengthen their activations. As a denoising step-level control method, our method can be combined with AAE directly by adding AAE gradient control in our generation algorithm process (both cross- and seld-attention control based on BoxNet in Algorithm~\ref{pseudo_code}).

\textbf{GLIGEN}~\cite{li2023gligen} achieves T2I generation with caption and bounding box condition inputs. Based on GLIGEN, we apply two-stage generation. In the first stage, given the prompt input, we use BoxNet to predict the box for each entity mentioned in the prompt. In the second stage, the predicted entity boxes and captions are fed into the GLIGEN model, and then attention mask control is adopted during generation to obtain layout-based images.

\subsection{Training and Evaluation Setup}
All the training details and hyper-parameter determination are presented in Appendix \ref{appendix:training_details}. For evaluation, we construct a new benchmark dataset to evaluate all methods with respect to semantic infidelity issues in T2I synthesis. To test the multi-object attribute binding capability of the T2I model, the input prompts should preferably consist of two or more objects with corresponding attributes (\textit{e.g.,} color). We come up with one unified template for text prompts: ``a [colorA][entityA] and a [colorB][entityB]'', where the words in square brackets will be replaced to construct the actual prompts. Note that [entity\#] can be replaced by an animal or an object word. We design two sets of optional vocabulary: the COCO category and the NON-COCO category (open domain). Every vocabulary contains 8 animals, 8 object items, and 11 colors, detailed in Appendix~\ref{appendix:evaluation_details}. For color-entity pairs in one prompt, we select colors randomly without repetition. For each prompt, we generate 60 images using the same 60 random seeds applied to all methods. For ease of evaluation, our prompts are constructed of color-entity pairs and the conjunction ``and''. Yet, our method is not limited to such patterns and can be applied to a variety of prompts with any type of subject, attribute, or conjunction.

\begin{table*}[tbp]
\centering
\caption{The quantitative evaluation results of five metrics for the six methods, including three baselines and three ablated variants of our method. Avg CLIP image-text/text-text and Min. Object Score measure multi-entity generation quality based on CLIP and the DINO score, respectively. Subj. Fidelity Score evaluates the correctness of entity and attribute generation through a user study. FID assesses the quality of generated images by measuring the feature distance between generated and real images.}
\resizebox{2.1\columnwidth}{!}{
\begin{tabular}{cccccccccc}
\toprule
\multirow{2}{*}{No.}  & \multirow{2}{*}{Model} & \multicolumn{2}{c}{Avg CLIP image-text} & \multirow{2}{*}{Avg CLIP text-text} &
\multicolumn{2}{c}{Min. Object Score} & \multicolumn{2}{c}{Subj. Fidelity Score} & FID\\ \cmidrule(l){3-4} \cmidrule(lr){6-9} 
 & & Full Prompt & Min. Object &  & COCO & NON-COCO & COCO & NON-COCO & COCO \\
\midrule
$[1]$ & STABLE~\cite{rombach2022high} & 0.3369 & 0.2414 & 0.7776 $\pm$ 0.0925 & 0.3973 $\pm$ 0.0021 & 0.3998 $\pm$ 0.0048 & 0.3021 $\pm$ 0.0759 & 0.3698 $\pm$ 0.0929 & 17.79 \\
 & StructureDiffusion~\cite{feng2022training} &  &  &  & 0.3728 $\pm$ 0.0038 & 0.3724 $\pm$ 0.0038 & 0.2767 $\pm$ 0.0566 & 0.3016 $\pm$ 0.0815 & - \\
 & AAE~\cite{chefer2023attend} & 0.3383 & 0.2437 & 0.7701 $\pm$ 0.0974 & 0.4438 $\pm$ 0.0027 & 0.4338 $\pm$ 0.0021 & 0.3552 $\pm$ 0.1043 & 0.3502 $\pm$ 0.0972 & - \\
\midrule
$[2]$ & $[1]$+BoxNet\&Cross Attn Mask Ctrl & - & - & - & 0.4010 $\pm$ 0.0028 & 0.4307 $\pm$ 0.0042 & - & - & - \\
$[3]$ & $[2]$+Uniq Mask & 0.3336 & 0.2414 & 0.7668 $\pm$ 0.0990 & 0.4456 $\pm$ 0.0039 & 0.4779 $\pm$ 0.0055 & 0.4141 $\pm$ 0.1087 & 0.3983 $\pm$ 0.1003 & 18.11 \\
$[4]$ & \textbf{$[3]$+Self Attn Ctrl(OURS)} & \textbf{0.3431} & \textbf{0.2516} & \textbf{0.7857 $\pm$ 0.1009} & \textbf{0.6028 $\pm$ 0.0047} & \textbf{0.5991 $\pm$ 0.0044} & \textbf{0.4331 $\pm$ 0.1404} & \textbf{0.4305 $\pm$ 0.1214} & \textbf{17.47} \\

\bottomrule
\end{tabular}\label{dino_bar_plot}
}
\end{table*}

\subsection{Qualitative Comparisons}
\begin{figure}[htbp]
	\centering
	\includegraphics[scale=0.34]{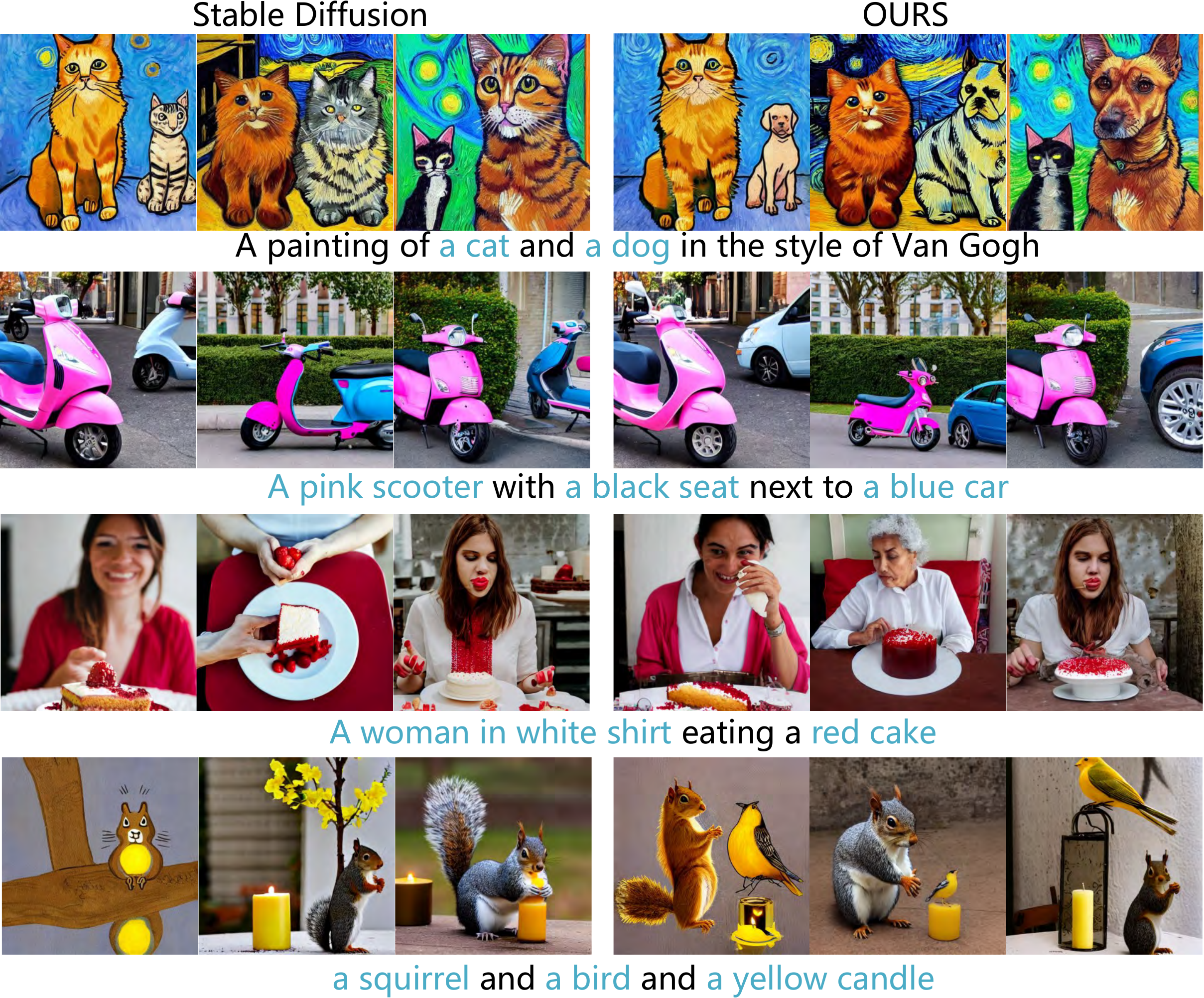}
	\caption{Comparison with complex prompts of more than two entities or multiple attributes. For each prompt, we apply the same set of random seeds on all methods. The entity-attribute pairs are highlighted in blue.}
	\label{fig:complex_results}
\end{figure}

In Fig.\ref{array_image_qualitative}, we present the generated results using fixed format self-built prompts as well as complex ones with more than two entities or intricate attributes (\textit{e.g.,} object actions, spatial relationships), which are taken from the AAE paper~\cite{chefer2023attend} and the test split of COCO datset~\cite{lin2014microsoft}. For each prompt, we show three images generated by the SD, StructureDiffusion, AAE, \textit{Ours} and \textit{Ours w/o Self-Attn Ctrl}, respectively. \textit{Ours} denotes the method with both cross- and self-attention mask control. As we can see, StructureDiffusion tends to generate images with missing entities and attribute leakage. For example, given ``a blue car and an orange bench'', its generated images may only contain an orange bench or an orange car that mixes the bench's color with the car's entity. As to AAE, its generated images still suffer from infidelity problems. Given ``a blue horse and a purple cake'', the AAE correctly generates the two mentioned entities in some cases but fails to bind each entity's color correctly (\textit{e.g.}, generating a purple horse or a white cake). In contrast, our method generates images that faithfully convey the semantics of the original prompt, showing robust attribute binding capability. This is because we explicitly enforce cross- and self-attention mask control over the attention areas to effectively alleviate attribute and entity leakage. For instance, the generated images of \textit{Ours} correctly correspond with the prompt ``a blue car and an orange bench'', where the colors of the car and bench do not leak or mix. Further more, Fig.\ref{fig:complex_results} and the last two columns of Fig.\ref{array_image_qualitative} show more comparisons with more than two entities and complex backgrounds, demonstrating its effectiveness when dealing with complicated prompts. Additionally, we provide more generation results based on simple or complex prompt descriptions in Appendix~\ref{appendix:c}.

\subsection{Quantitative Analysis}
Firstly, we quantify the performance of every competing approach through Average CLIP image-text similarities and Average CLIP text-text similarities from AAE~\cite{chefer2023attend}. But since global CLIP image-text scores are insensitive to entity missing and attribute leaking issues, we further propose to use Grounding DINO score~\cite{liu2023grounding} as a more fine-grained evaluation metric which focus on local object level. However, even Grounding DINO score takes into account issues of entity missing and entity leakage, it's still insensitive to entity attributes so that it does not reflect whether attributes such as color are generated correctly or not. To measure the overall generation performance of both entities and attributes, taking full account of the three infidelity issues, we futher conduct a user study. Additionally, we use FID~\cite{heusel2017gans} to assess the overall quality of generated images on 10k samples of the COCO dataset by calculating the distance between feature vectors of generated and real images. All detailed analysis and descriptions of the evaluation metrics (both objective and subjective) are presented and discussed in Appendix~\ref{appendix:evaluation_metrics}.

\textbf{DINO Similarity Scores.} Grounding DINO is an open-set object detection model, which accepts an image-text pair and predicts object boxes. Each predicted object box has similarity score ranging from $0$ to $1$ across all input words. We use the DINO score for the most neglected entity as the quantitative measure of multi-entity generation performance. To this end, we compute the DINO score between every entities exist in the original prompt of each generated image. Specifically, given the prompt ``a [colorA] [EntityA] and a [colorB] [EntityB]'', we extract the names of the entities (\textit{e.g.,} ``a [EntityA]'' and ``a [EntityB]''), and feed them with the generated image into the DINO model to obtain boxes and corresponding similarity scores. If one entity has multiple detected boxes, we adopt the highest similarity score across all boxes as its score. Conversely, if one entity has no detected boxes, we assign a score of $0$ to it. Given all the entity scores (two in our case) for each image, we are more concerned with the smallest one as this would correspond to the issues of entity missing and entity leakage. The average of the smallest DINO scores across all seeds and prompts is taken as the final metric of each method, called \emph{Minimum Object Score}. 
\begin{table}[tbh]
\centering
\caption{Comparison of the Min. Object Scores for the proposed plugin solutions, split by evaluation datasets. The first column indicates different states of methods. We show the performance of the three methods after being plugged with our proposed techniques, respectively.}
\resizebox{1.0\columnwidth}{!}{
\begin{tabular}{c|ccc}
\hline
\multirow{2}{*}{State} & \multicolumn{3}{c}{COCO} \\ \cline{2-4} 
 & STABLE & AAE & GLIGEN \\ \hline
BASE & 0.3973 $\pm$ 0.0021 & 0.4438 $\pm$ 0.0027 & 0.5046 $\pm$ 0.0022 \\
BoxNet & - & - & 0.5788 $\pm$ 0.0005 \\
w/ Cross-Attn Ctrl & 0.4456 $\pm$ 0.0039 & 0.4831 $\pm$ 0.0033 & 0.6200 $\pm$ 0.0010 \\
\textbf{w/ Cross- and Self-Attn Ctrl} & \textbf{0.6028 $\pm$ 0.0047} & \textbf{0.6257 $\pm$ 0.0056} & \textbf{0.6718 $\pm$ 0.0045} \\ \hline \hline
 & \multicolumn{3}{c}{NON-COCO} \\ \hline
BASE & 0.3998 $\pm$ 0.0048 & 0.4338 $\pm$ 0.0021 & 0.4574 $\pm$ 0.0059 \\
BoxNet & - & - & 0.5585 $\pm$ 0.0023 \\
w/ Cross-Attn Ctrl & 0.4779 $\pm$ 0.0055 & 0.4957 $\pm$ 0.0018 & 0.6330 $\pm$ 0.0013 \\
\textbf{w/ Cross- and Self-Attn Ctrl} & \textbf{0.5991 $\pm$ 0.0044} & \textbf{0.5918 $\pm$ 0.0028} & \textbf{0.6839 $\pm$ 0.0024} \\ \hline

\end{tabular}\label{line_plot_cropped}
}
\end{table}

\textbf{User Study.} We also perform a user study to analyze the fidelity of the generated images. 25 prompts on COCO or NON-COCO datasets are randomly sampled to generate 10 images, while each method shares the same set of random seeds. For the results of each prompt ``a [colorA] [EntityA] and a [colorB] [EntityB]'', we ask the respondents to answer two questions: (1) ``is there a [colorA] [EntityA] in this picture?'' and (2) ``is there a [colorB] [EntityB] in this picture?''. An answer of ``YES'' indicates both the color and entity can match the given text prompt. Only if the answer to both questions is yes, can this generated image be considered correct. We obtain \textit{Subjective Fidelity Score} by counting the correct proportion of all 25$\times$10 images on COCO or NON-COCO datasets.

\textbf{Comparison to Prior Work.} The quantitative results on the COCO and NON-COCO datasets are summarized in Table~\ref{dino_bar_plot}. We compare our method with three baselines (STABLE, AAE, Structure) in terms of five metrics.
We have replicated the test dataset and metrics used in \cite{chefer2023attend}, which are recorded in Table ~\ref{dino_bar_plot} as the Average CLIP image-text similarities and Average CLIP text-text similarities. Further, the Min. Object Score, Subj. Fidelity Score, and FID distance are calculated for better comparison. As shown, our method consistently outperforms all competing methods, with significant improvements in the fidelity of multi-entity generation and the correctness of attribute bindings between colors and entities. StructureDiffusion obtains scores similar to those of SD (even slightly lower), which is consistent with~\cite{chefer2023attend}. And AAE gains scores slightly higher than SD. Although trained on the COCO dataset, our method still performs well in the NON-COCO (open-domain) dataset, exhibiting good generalization ability. Additionally, our method achieves a slightly better FID than SD, indicating that the generation quality does not decrease after applying our attention mask control strategy.


\textbf{Ablation Study.} 
For the ablation study, we start with the original SD model and gradually add constitutive elements until we reach the complete OURS method. Whereas $[1]$ represents the SD model, $[2]$ applies BoxNet and non-uniq cross-attention mask control and can obtain experimental results that are comparable to (slightly better than) those of $[1]$. $[3]$ applies uniq mask control based on $[2]$, and can achieve similar metric results to AAE. By finally adding the self-attention control, we have the OURS method, marked as $[4]$.Table~\ref{dino_bar_plot} shows the contribution of different components of our model to the compositional T2I synthesis.

\subsection{Plugin Experiments}
In this section, we verify the effectiveness of our proposed two plugin solutions by comparing the results of existing models (AAE and GLIGEN) with and without our method. The experiment results are shown in Table~\ref{line_plot_cropped}. The first column indicates different states of methods. The \textbf{BASE} indicates the original state of each method as described in their papers. Note that in this state, we randomly generate object boxes as additional input conditions for GLIGEN. In the \textbf{BoxNet} state, the predicted boxes of BoxNet are used to replace the input random boxes for GLIGEN, while the remaining two states represent the results after imposing our attention mask control strategy on the three methods. As we can see, the generation quality of AAE and GLIGEN is significantly improved after being plugged into our strategy. Both the cross- and self-attention control can alleviate the infidelity issues, while the self-attention control contributes more to the improvement of Min. Object Score. However, in the open-domain NON-COCO evaluation, \textit{AAE w/ Cross- and Self-Attn Ctrl} unexpectedly perform worse than its counterpart in SD. We suspect that this is because the predicted boxes of the BoxNet on the NON-COCO dataset do not overlap with the region of interest in AAE, resulting in a conflict between these two methods. More qualitative results can be found in Appendix~\ref{appendix:c}.

\section{Conclusion}

In this paper, we present a novel attention mask control strategy based on the proposed BoxNet. We first train a BoxNet to predict object boxes when given the noisy image, timestep and text prompt as input. We then enforce unique mask control over the cross- and self-attention maps based on the predicted boxes, through which we alleviate three common issues in the current Stable Diffusion: attribute leakage, entity leakage, and missing entities. During the whole training process of BoxNet, the parameters of diffusion model are frozen. Our method guides the diffusion inference process on the fly, which means it can be easily incorporated into other existing diffusion-based generators when given a trained BoxNet.

\bibliography{main}

\clearpage
\newpage
\section*{Appendix}
\appendix
\ In this supplementary, we first detailedly describe the training and evaluation settings, including datasets, haper-parameters and evaluation benchmarks in Appendix \ref{appendix:method_details}. Then, in Appendix \ref{appendix:evaluation_metrics}, we discuss different evaluation metrics and analyze the metric chosen logic of our method. Finally, we present more visualization results to further compare our approach to other SOTA methods, to demonstrate the effectiveness of our method as a plugin and to show our limitations as well in Appendix \ref{appendix:visualizations}.
\section{Method Details}
\label{appendix:method_details}

\subsection{Self Attention Mask Control}
\label{appendix:self-attention_mask_control}
As Fig.\ref{self_attn_ctrl} in Appendix shows, we provide a detailed illustration of the Self Attention Mask Control method.

\begin{figure}[htbp]
	\centering
	\includegraphics[scale=0.25]{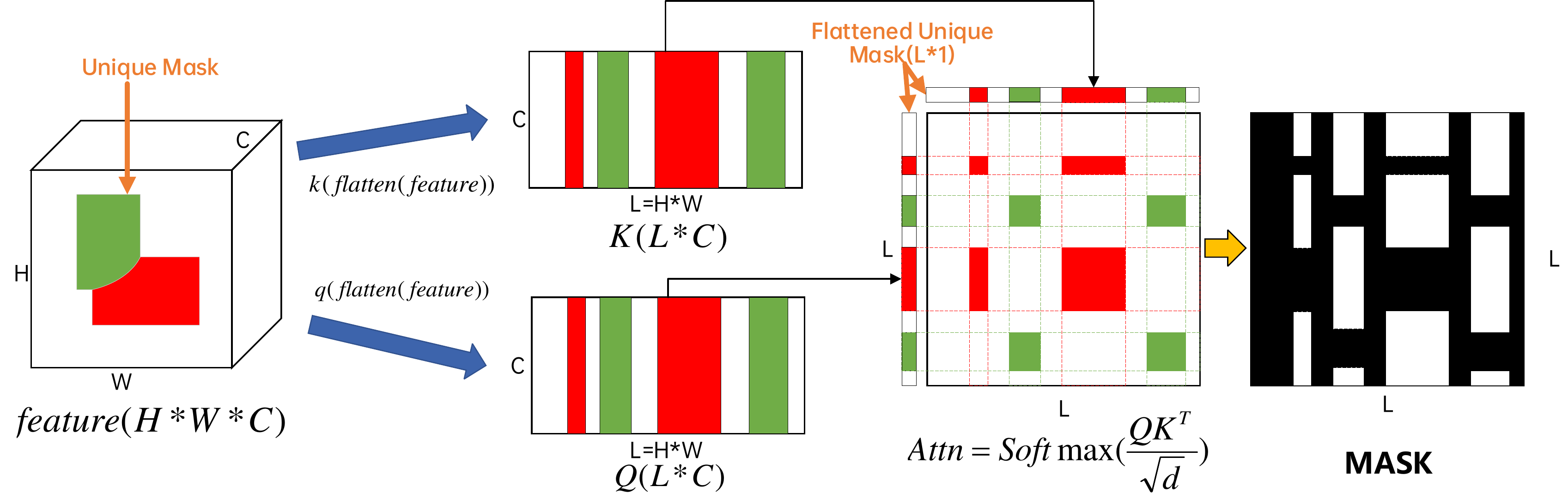}
	\caption{A detailed explanation of Self Attention Control: By using the Unique Mask to restrict the $L*L$ attention map, the generated entities are guided to focus on the bounding box regions they occupy.}
	\label{self_attn_ctrl}
\end{figure}

\subsection{Training Details}
\label{appendix:training_details}
To train the BoxNet to predict entity boxes, we use the images along with its bounding boxes with 80 object categories and captioning annotations from the COCO (Common Objects in Context) 2014 dataset~\cite{lin2014microsoft}, which consists of 83K training images and 41k validation images. Each image is annotated with bounding boxes and 5 captions. In all experiments, we adopt the Stable Diffusion V-1.5 checkpoint\footnote{\url{https://huggingface.co/runwayml/stable-diffusion-v1-5}} as the model base for a fair comparison. The parameters of diffusion model are frozen during the whole training process of the BoxNet. The BoxNet is a transformer-based architecture with 6 encoder and 6 decoder layers~\cite{vaswani2017attention}. For the initialization of BoxNet, we use the Xavier init. We use AdamW optimizer to train the BoxNet for 150k steps on 8*A100 with parameters $lr$ = 0.0004, $weight\_decay=0.0001$, $warmup\_steps=10k$. 
For those hyper-parameters, we set transformer decoder max sequence length $M$ to 30, penalty of class mismatch $\lambda$ to 100 and loss weights $\lambda_{iou} = 2$, $\lambda_{L1} = 5$.

\subsection{Evaluation Details}

\label{appendix:evaluation_details}
\textbf{Benchmark.} In order to fairly compare different existing methods with our method, we construct a benchmark evaluation dataset based on~\cite{chefer2023attend}. The difference is that we abandon the distinction between object items and animals, freely combine the two as a collection of entities, and assign attributes (colors) to all the entities at the same time. In addition, since our BoxNet is trained on the COCO dataset, in order to verify the generalization ability of our model, we design two data categories for comparison. The object items and animals in the COCO category are drawn from the COCO dataset \cite{lin2014microsoft}, whereas those in the NON-COCO category are drawn from sources other than the COCO dataset.  Both categories share the same color collection. 
When creating a prompt, the entity collection will be comprised of the object item and animal collections. We have 8 animals and 8 object items in each category, for a total of 16 entities, and we compose each two different entities using the evaluation prompt template to generate 120 text prompts.
Furthermore, when creating the evaluation prompt, we assign different colors to all of the entities at random to observe the problem of attribute leakage. Table \ref{eval_category} shows the detailed categories of our evaluation dataset.
During the evaluation phase, all T2I synthesis methods will generate images using the same 60 random seeds based on each text prompt.

\begin{table}[h]
\centering
\caption{Evaluation Datasets. We list the animals, objects, and colors used to define two evaluation data subsets for COCO category and NON-COCO category, respectively.}
\resizebox{1.0\columnwidth}{!}{
\begin{tabular}{c|c|l}
\hline
                                                                          & placeholder & \multicolumn{1}{c}{vocabulary}                                                                                      \\ \hline
                                                                          & animals     & \begin{tabular}[c]{@{}l@{}}cat, dog, bird, bear, horse, elephant,\\ sheep, giraffe\end{tabular}                     \\ \cline{2-3} 
COCO category                                                             & object items     & \begin{tabular}[c]{@{}l@{}}backpack, suitcase, chair, car, couch, bench,\\ cake, umbrella\end{tabular}              \\ \cline{2-3} 
                                                                          & colors      & \begin{tabular}[c]{@{}l@{}}red, orange, yellow, green, blue, purple,\\ pink, brown, gray, black, white\end{tabular} \\ \hline
                                                                          & animals     & \begin{tabular}[c]{@{}l@{}}tiger, panda, lion, fox, squirrel, turkey,\\ penguin, turtle\end{tabular}                \\ \cline{2-3} 
\begin{tabular}[c]{@{}c@{}}NON-COCO category\\ (open domain)\end{tabular} & object items     & \begin{tabular}[c]{@{}l@{}}shoes, television, watermelon, candle, bucket,\\ hammock, pumpkin, carrot\end{tabular}   \\ \cline{2-3} 
                                                                          & colors      & \begin{tabular}[c]{@{}l@{}}red, orange, yellow, green, blue, purple,\\ pink, brown, gray, black, white\end{tabular} \\ \hline
\end{tabular}
}
\label{eval_category}
\end{table}

\begin{figure}[htbp]
	\centering
	\includegraphics[scale=0.23]{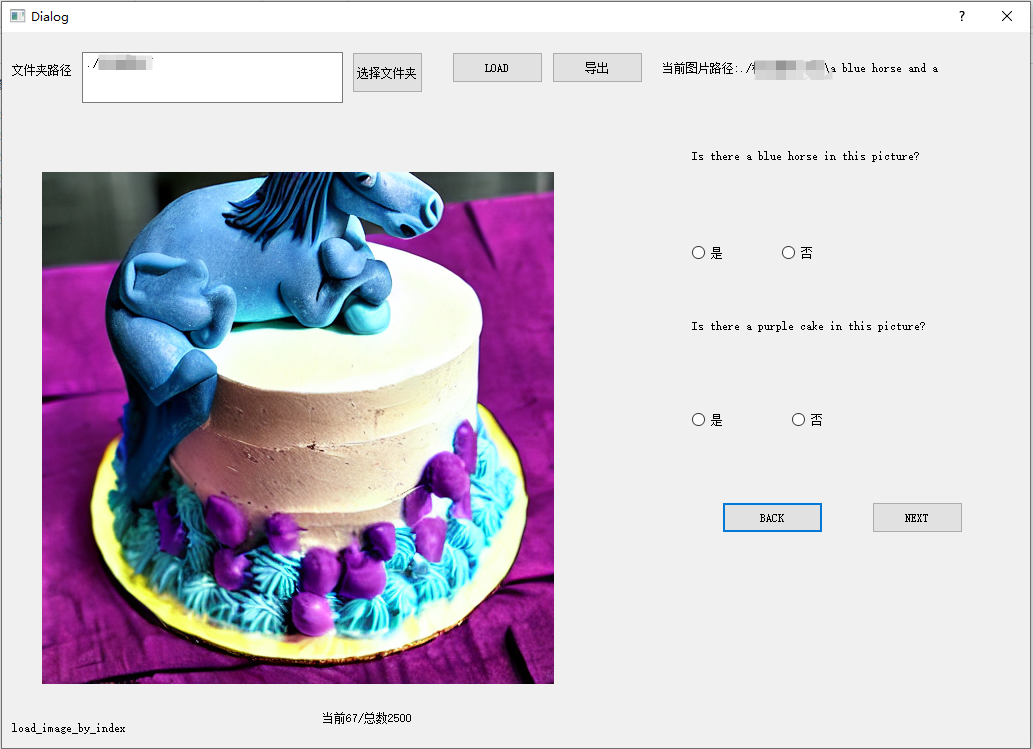}
	\caption{The UI interface of our image annotation tool is designed for users to sequentially answer two yes-or-no questions for each image.}
	\label{ui}
\end{figure}

\textbf{User Study.} In our user study experiment, we recruited 11 respondents to assess each image and answer two questions ("Is there a [colorA] [EntityA] in this picture?" and "Is there a [colorB] [EntityB] in this picture?"). We designed a simple annotation tool UI as shown in Fig.\ref{ui}.

\section{Evaluation Metrics}
\label{appendix:evaluation_metrics}

\subsection{FID Score}
We use the FID metric on the COCO dataset to assess the image quality produced by various methods. We randomly sample 10k text prompts from the COCO validation dataset and use the same random seeds to generate the same number of images and calculate the FID score. Our method differs from stable diffusion in that the input prompt must be parsed. To extract the description of entities with attributes, we use the open-source text parsing tool mentioned in \cite{feng2022training}.
However, we have discovered that there are significant errors in the entity descriptions extracted in this manner, which has a negative impact on the generation quality. As a result, we filter the extracted entity descriptions further by creating a vocabulary of entity words and using simple keyword filtering, as shown in Table~\ref{filtered_prompt}.

\begin{table}[!htbp]
\centering
\caption{We show some examples of parsing result filtering, including text spans extracted with an open-source parsing tool and filtered text spans. Our simple filtering rules can remove some incorrect spans from the generated results. BoxNet is used to obtain corresponding boxes from the filtered text spans, and attention mask control is used to control image generation.}
\resizebox{1.0\columnwidth}{!}{
\begin{tabular}{c|c|c} 
\hline
 & \textbf{Open-source Tool} & \textbf{Filtered} \\
\hline
\multirow{2}*{Example 0} & \multicolumn{2}{c}{\textit{Prompt:} a white clock tower with a clock on each of it's sides} \\ 
\cline{2-3}
 & \begin{tabular}[c]{@{}c@{}}"a white clock tower", \\"a clock", "it 's"\end{tabular} & \begin{tabular}[c]{@{}c@{}}"a white clock tower", \\"a clock"\end{tabular} \\ 
\hline
\multirow{2}*{Example 1} & \multicolumn{2}{c}{\textit{Prompt:} a man is sitting on the back of an elephant} \\ 
\cline{2-3}
 & \begin{tabular}[c]{@{}c@{}}"a man", "the back", \\"an elephant"\end{tabular} & \begin{tabular}[c]{@{}c@{}}"a man", \\"an elephant"\end{tabular} \\ 
\hline
\multirow{2}*{Example 2} & \multicolumn{2}{c}{\textit{Prompt:} many different fruits are next to each other} \\ 
\cline{2-3}
 & \begin{tabular}[c]{@{}c@{}}"many different fruits", \\"each other"\end{tabular} & \begin{tabular}[c]{@{}c@{}}"many different fruits"\end{tabular} \\ 
 \hline
\multirow{2}*{Example 3} & \multicolumn{2}{c}{\textit{Prompt:} a large red umbrella with other colors around the center pole} \\ 
\cline{2-3}
 & \begin{tabular}[c]{@{}c@{}}"a large red umbrella", \\"other colors", \\"the center pole"\end{tabular} & \begin{tabular}[c]{@{}c@{}}"a large red umbrella"\end{tabular} \\ 

\hline
\end{tabular}
}
\label{filtered_prompt}
\end{table}

\subsection{DINO Score}
\begin{figure}[htbp]
	\centering
	\includegraphics[scale=0.30]{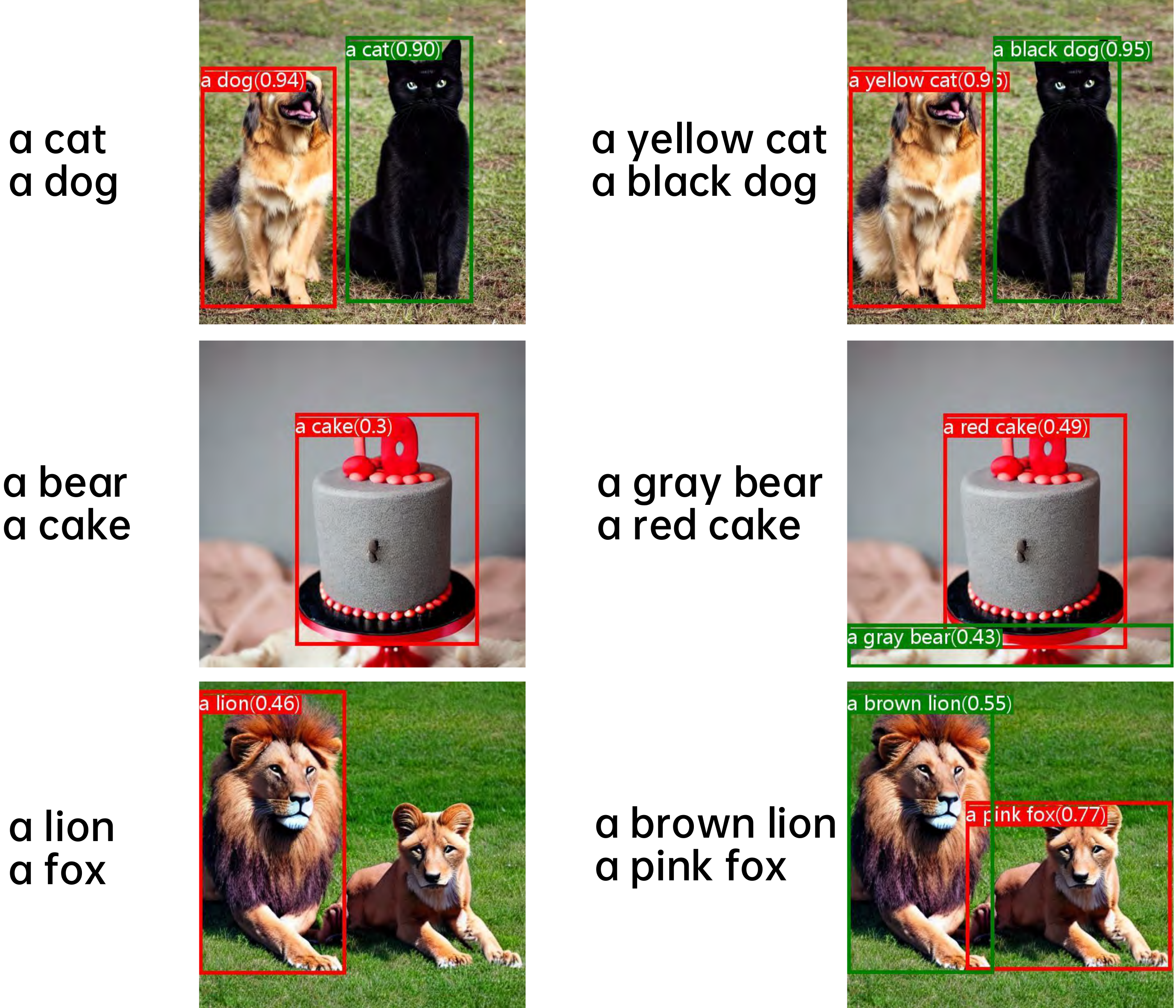}
	\caption{Here are several bad detection results of Grounding DINO model when the input prompts contain entities with attributes. We conducted two experiments for comparison: the left side remains consistent with the main text by using the prompt ``a [entity*]'' for detection and keeping only one box with highest score for each entity; the right side presents obvious entity confusion and false positive detection by using the prompt ``a [color*] [entity*]'' with attributes for detection.}
	\label{dino_badcase}
\end{figure}

The DINO score is the primary quantitative metric in our study, and it is based on the Grounding DINO model for open-domain object detection. The Grounding DINO model detects target objects with consistent accuracy. When detecting multiple objects with different attributes in the same image, however, false detections can occur. As shown in Fig.\ref{dino_badcase}, object detection using entity names as prompts is generally correct, but using entities with attributes as prompts increases the likelihood of false detections, especially when the input generated images are problematic. Attribute words (colors) can easily lead the model astray and cause it to locate the incorrect entity. As a result, we only use entity words as input to detect objects and evaluate all models' ability to generate entities. As for attribute evaluation, it will be completed through the user study.

\subsection{CLIP Score}
\begin{figure}[htbp]
	\centering
	\includegraphics[scale=0.4]{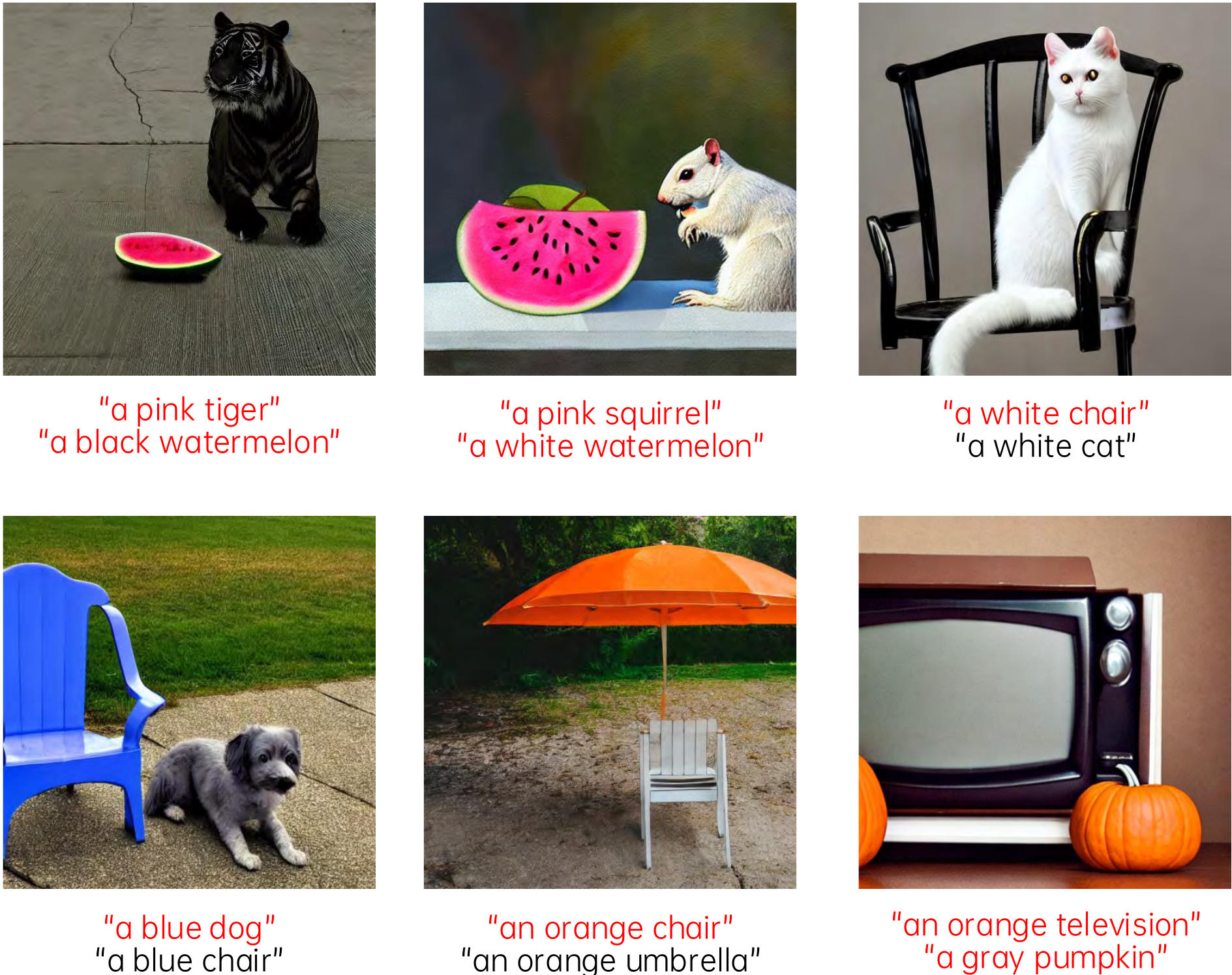}
	\caption{For some badcases of the CLIP score, we list two entity prompts with the highest scores for each image. If the color of an entity prompt does not match that of the entity in the image prompt, we highlight it in red.}
	\label{clip_badcase}
\end{figure}

As a common evaluation metric in T2I (text-to-image) generation papers, we initially considered using CLIP (Contrastive Language-Image Pre-training) \cite{radford2021learning} for model evaluation. However, we discovered that CLIP has poor color discrimination and thus struggles to judge the correctness of entity attributes. To test this, we randomly selected 100 images that were correctly identified by all respondents during the user study and calculated the CLIP score for each entity based on the text prompts ``a [colorA][entityA]'' and ``a [colorB][entityB]''. The method involves replacing the color in the entity prompt with all of the colors from the color set in the test dataset and then using CLIP to calculate the image's score over entity prompts with all of the different colors. We considered it a correct judgment only when the score on the entity prompt with the correct color is the highest. We calculated the correctness of 200 entities across all 100 images and discovered an average correctness rate of only 43$\%$. Fig.\ref{clip_badcase} depicts some CLIP score failures. Although there are the aforementioned issues with the CLIP Score, we still incorporated this metric into our testing as a reference. In Table ~\ref{dino_bar_plot} of the main text, we applied two quantitative metrics, namely Avg CLIP image-text and Avg CLIP text-text, which are reproductions of the metrics used in the study by \cite{chefer2023attend}. These metrics do not provide a good level of discriminability for various methods, although our approach still achieved state-of-the-art (SOTA) performance.

\section{Additional Qualitative Results}
\label{appendix:visualizations}
In this section, we provide additional visualization results and comparisons.

\begin{figure*}[htbp]
	\centering
	\includegraphics[scale=0.30]{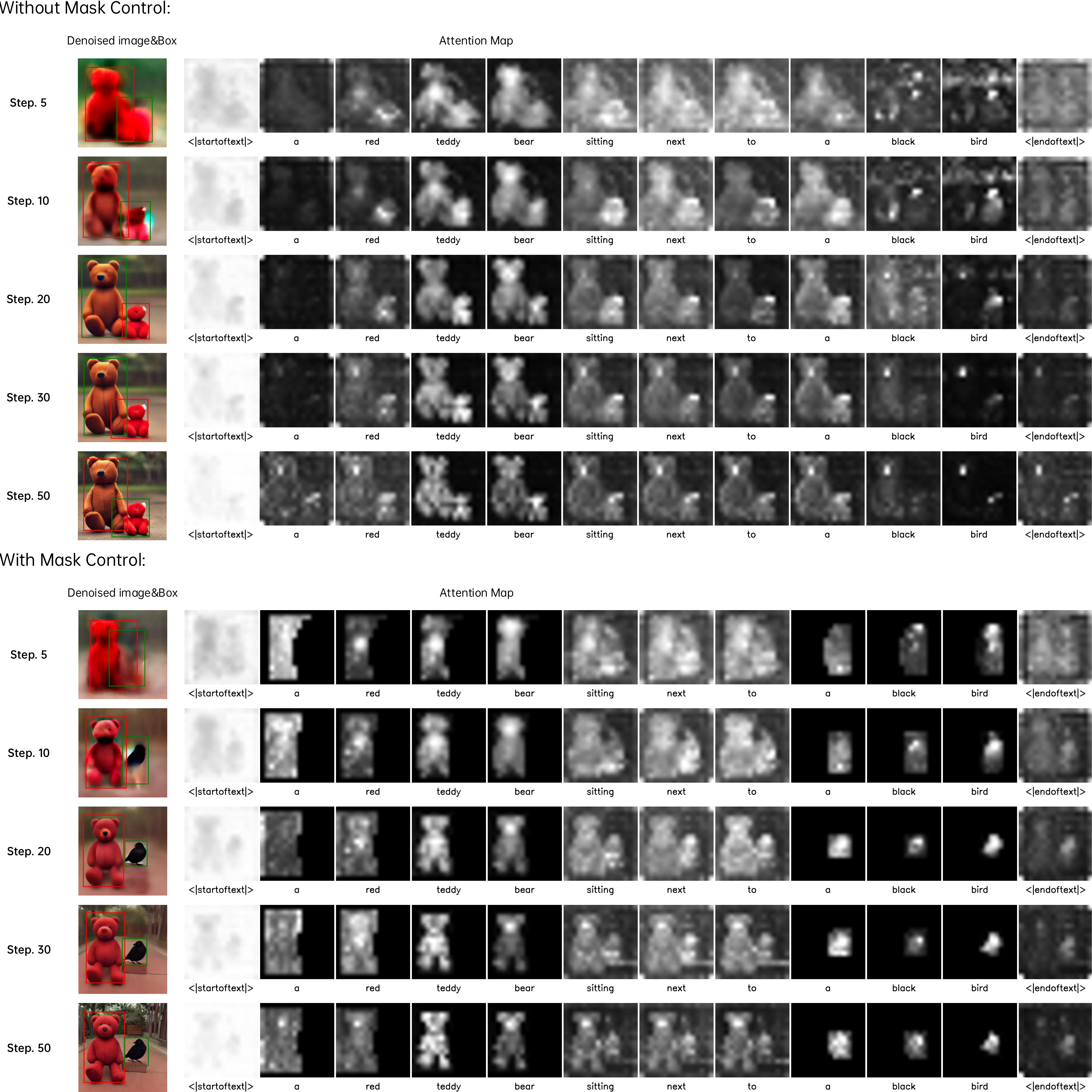}
	\caption{Attention map and BoxNet predicts on different sampling steps. This shows the results with and without Attention Mask Control. The 1st example illustrates how our method aids in optimizing the generated outcomes in the presence of attribute leakage in the SD model.}
	\label{additional_results_compre_1}
\end{figure*}

\begin{figure*}[htbp]
	\centering
	\includegraphics[scale=0.50]{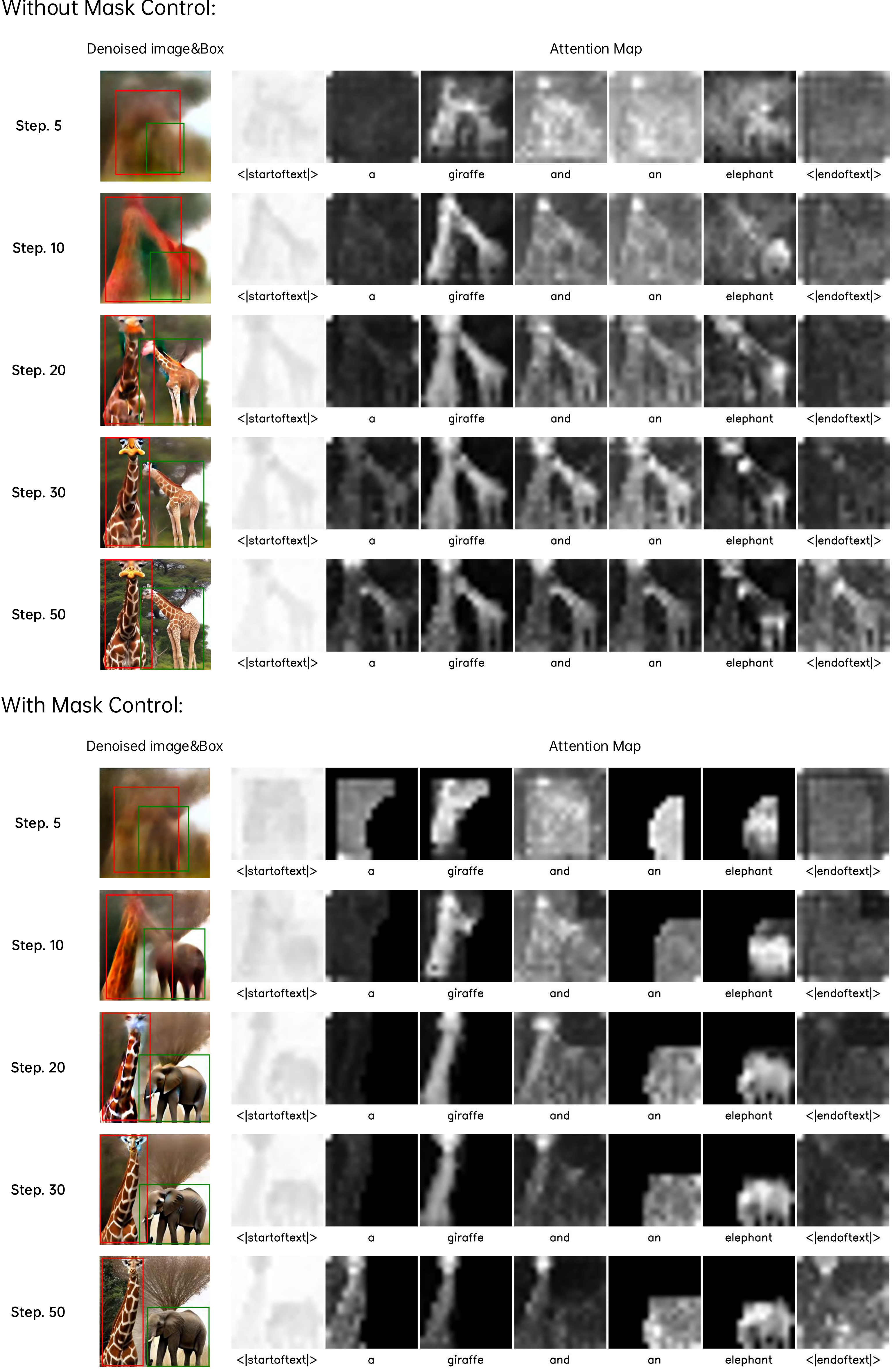}
	\caption{Attention map and BoxNet predicts on different sampling steps. This shows the results with and without Attention Mask Control. The 1st example illustrates how our method aids in optimizing the generated outcomes in the presence of entity leakage in the SD model.}
	\label{additional_results_compre_2}
\end{figure*}

\begin{figure*}[htbp]
	\centering
	\includegraphics[scale=0.40]{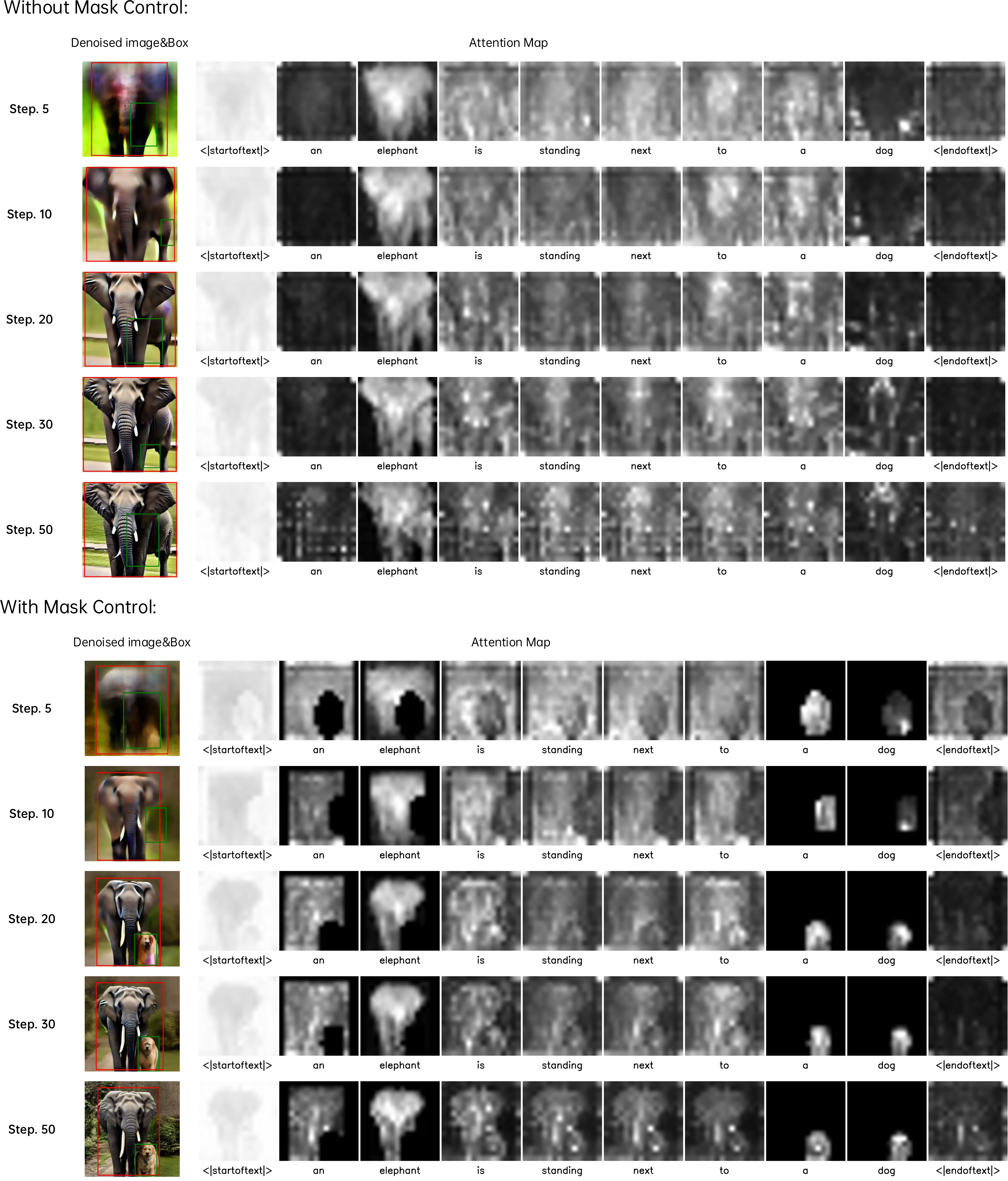}
	\caption{Attention map and BoxNet predicts on different sampling steps. This shows the results with and without Attention Mask Control. The 1st example illustrates how our method aids in optimizing the generated outcomes in the presence of missing entities in the SD model.}
	\label{additional_results_compre_3}
\end{figure*}

\begin{figure*}[htbp]
	\centering
	\includegraphics[scale=0.40]{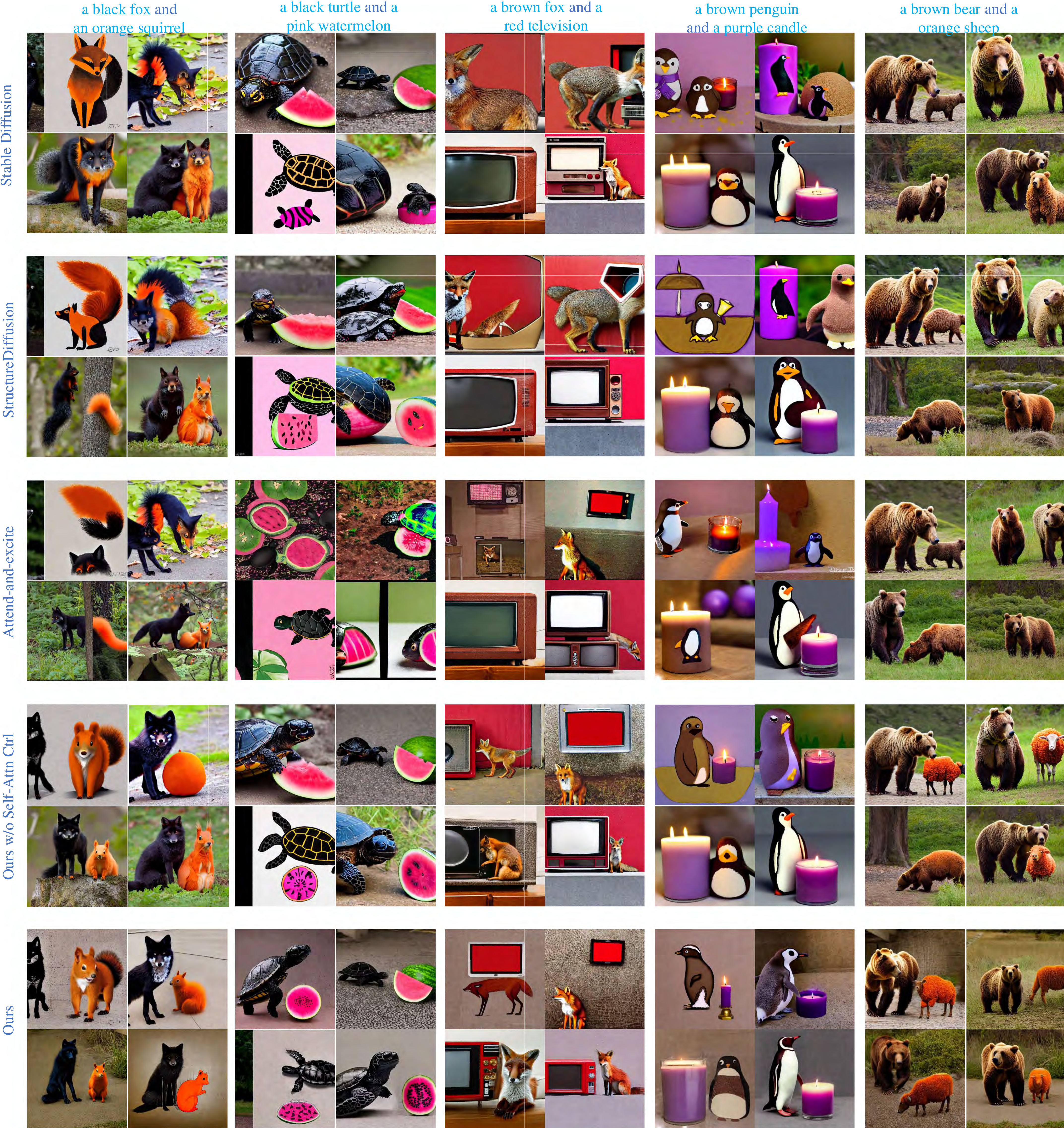}
	\caption{Additional results on our benchmark evaluation dataset. For each prompt, we apply the same set of random seeds on all methods. The entity-attribute pairs are highlighted in blue.}
	\label{additional_results_1}
\end{figure*}

\begin{figure*}[htbp]
	\centering
	\includegraphics[scale=0.5]{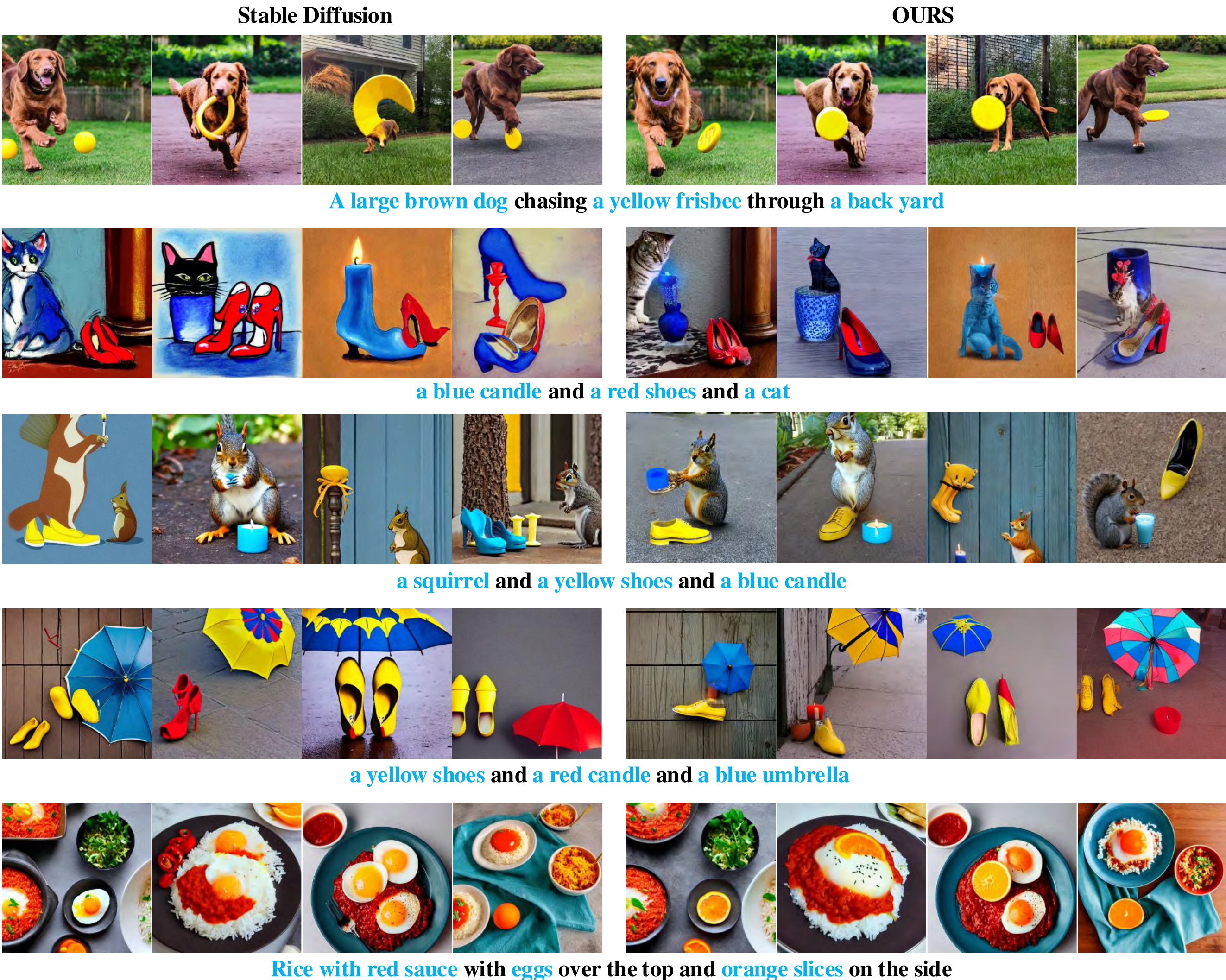}
	\caption{Comparison with complex prompts of more than two entities or multiple attributes. For each prompt, we apply the same set of random seeds on all methods. The entity-attribute pairs are highlighted in blue.}
	\label{array_image_qualitative_2}
\end{figure*}

\begin{figure*}[htbp]
	\centering
	\includegraphics[scale=0.36]{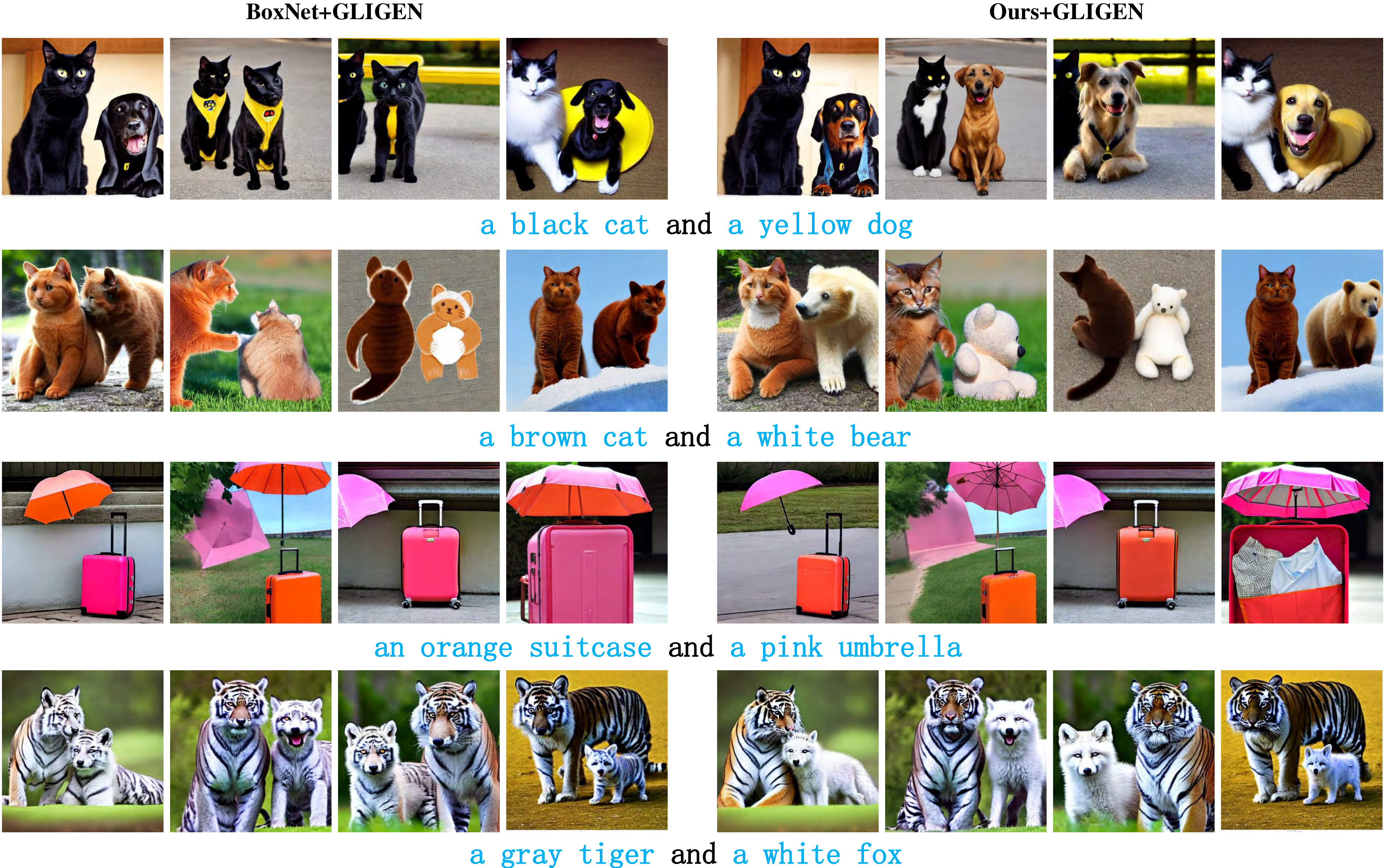}
	\caption{comparison of our method as a GLIGEN plugin. For each prompt, we apply the same set of random seeds on all methods. The entity-attribute pairs are highlighted in blue}
	\label{additional_results_3}
\end{figure*}

\begin{figure*}[htbp]
	\centering
	\includegraphics[scale=0.38]{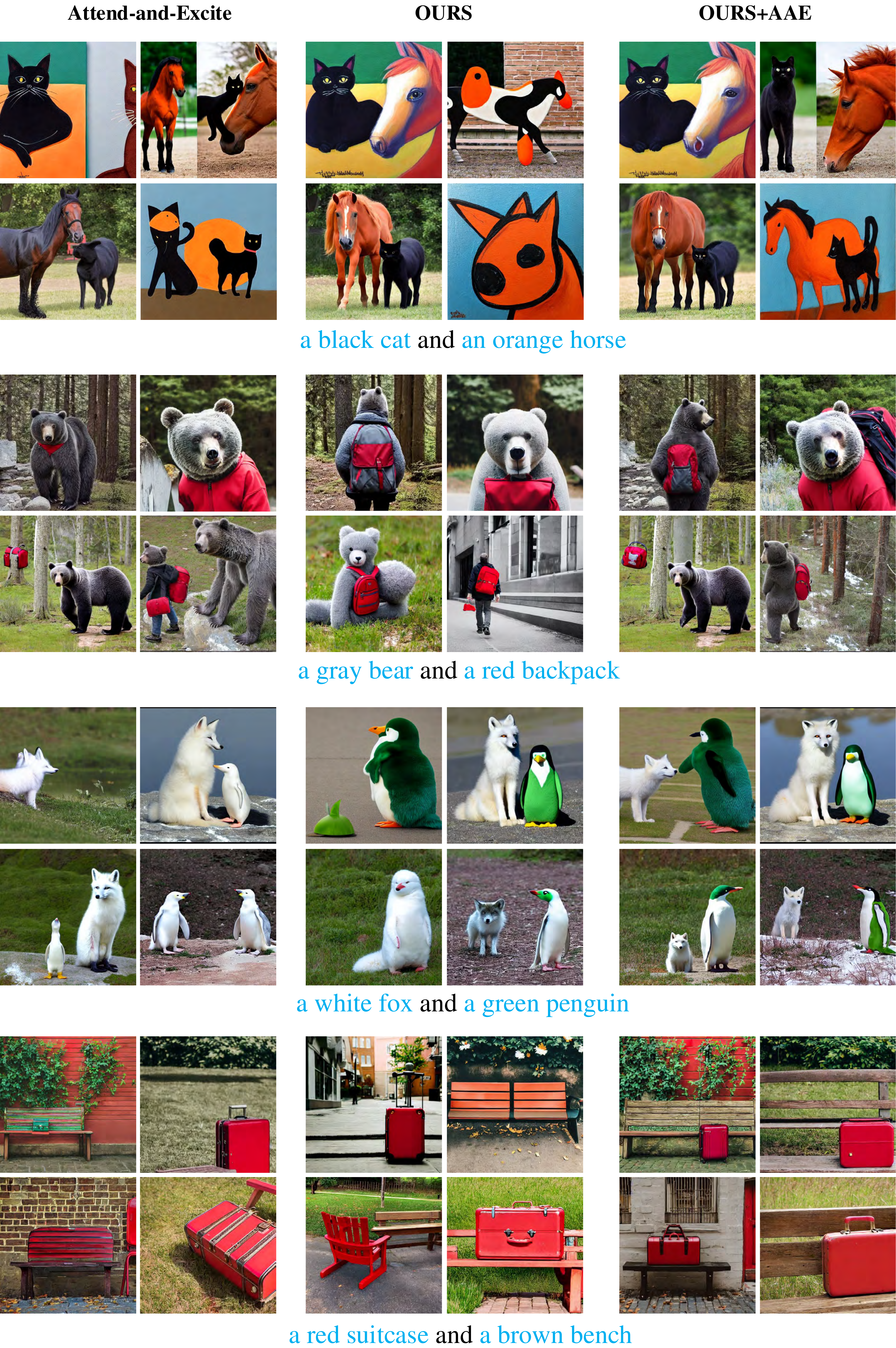}
	\caption{comparison of our method as an AAE plugin. For each prompt, we apply the same set of random seeds on all methods. The entity-attribute pairs are highlighted in blue}
	\label{additional_results_4}
\end{figure*}

\begin{figure*}[htbp]
	\centering
	\includegraphics[scale=0.36]{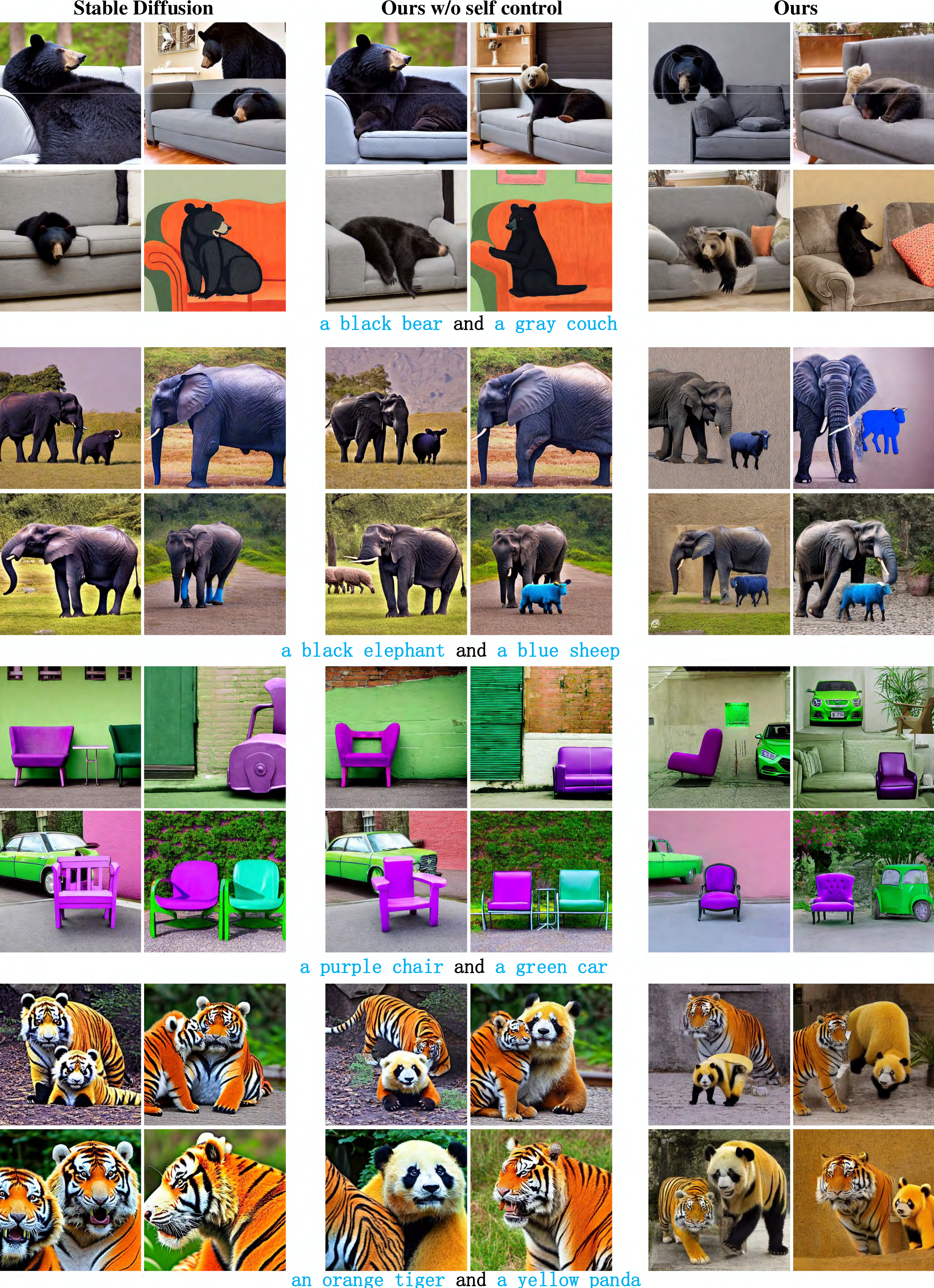}
	\caption{We select some examples of images generated by our method with degraded quality. We compared the original stable diffusion model, OURS w/o self attention control, and OURS, with all generated images using the same random seed.}
	\label{additional_results_5}
\end{figure*}

\begin{itemize}
  \item Fig. \ref{additional_results_compre_1}, Fig. \ref{additional_results_compre_2} and Fig. \ref{additional_results_compre_3} presents supplementary results for Fig.\ref{box} in the main text, illustrating the predictive outcomes of BoxNet as well as the attention maps before and after the application of Attention Mask Control. The three examples correspond to the three issues identified in the main text regarding the SD model, namely attribute leakage, entity leakage, and missing entities. By incorporating our approach, we successfully mitigated these three issues and generated images containing accurate entities and attributes.
  \item Fig. \ref{additional_results_1} shows additional results on our self-built evaluation dataset of several comparable approaches, including Stable Diffusion \cite{rombach2022high}, SructureDiffusion \cite{feng2022training} and Attend-and-Excite \cite{chefer2023attend}, whereas Fig. \ref{array_image_qualitative_2} shows examples generated based on some realistic complex prompts.
  \item Fig. \ref{additional_results_3} and Fig. \ref{additional_results_4} show the qualitative results of our method as a plugin for the AAE and GLIGEN methods.
  \item Fig. \ref{additional_results_5} illustrates the limitations of our approach. Although our method does not result in a decrease in FID score, there may be instances where image quality suffers slightly during the generation of multi-entity images. This degradation may appear as an unnatural integration of entities and backgrounds, or as a "tearing" phenomenon in the generated background. If, on the other hand, we do not use self-attention control, the generated results are comparable to those of the SD model and do not exhibit this drop in quality, even if the generated entities and attributes may not remain correct.
\end{itemize}

\label{appendix:c}

\end{document}